\documentclass[a4paper, 1p, oneside, onecolumn, preprint]{elsarticle}
\usepackage{lineno}
\modulolinenumbers[5]
\usepackage{xcolor}
\usepackage{lipsum}
\usepackage{multirow}
\usepackage{amsmath}
\usepackage{empheq}
\usepackage{textcomp}
\usepackage{indentfirst}
\usepackage{amssymb}
\usepackage{algorithm}
\usepackage{algorithmicx}
\usepackage{algpseudocode}
\usepackage{graphicx}
\usepackage{tabulary,booktabs}
\usepackage{amsthm}
\usepackage{bbm}

\newtheorem{thm}{Theorem}
\newtheorem{lemma}{Lemma}

\usepackage{subfigure}
\usepackage{bm}
\usepackage{url}

\newcolumntype{M}[1]{>{\centering\arraybackslash}m{#1}}

\def\R{\mathbb{R}}
\def\E{\mathbb{E}}
\def\P{\mathbb{P}}

\def\N{\mathcal{N}}
\def\x{\mathbf{x}}

\def\w{\mathbf{w}}
\def\X{\mathcal{X}}
\def\L{{L}}
\def\Y{\mathcal{Y}}

\def\D{\mathbf{D}}
\def\W{\mathbf{W}}
\def\H{\mathbf{H}}

\def\M{\mathbf{M}}

\def\z{\mathbf{z}}
\def\v{\mathbf{v}}
\def\a{\mathbf{a}}
\def\u{\mathbf{u}}
\def\I{\mathbf{I}}

\def\M{\mathbf{M}}
\def\P{\mathbf{P}}
\def\U{\mathbf{U}}
\def\Q{\mathbf{Q}}
\def\thetab{\bm{\theta}}

\def\H{\bm{H}}
\def\xib{\bm{\xi}}
\def\phib{\bm{\phi}}
\def\psib{\bm{\psi}}
\def\mub{\bm{\mu}}
\newcommand{\RNum}[1]{\uppercase\expandafter{\romannumeral #1\relax}}

\begin{document}

\title{Provable Convergence of Nesterov's Accelerated Gradient Method for Over-Parameterized Neural Networks}
\author[1]{Xin Liu}\ead{1036870846@qq.com}
\author[1]{Zhisong Pan\corref{cor1}}\ead{hotpzs@hotmail.com}
\author[2]{Wei Tao} \ead{wtao_plaust@163.com}
\tnotetext[t1]{This work was supported by National Natural Science Foundation of China (No.62076251 and No.62106281).}
\cortext[cor1]{Corresponding author}

\affiliation[1]{organization={Command $\&$ Control Engineering College, Army Engineering University of PLA}, 
                 postcode={210007}, 
                 city={Nanjing}, 
                 country={P.R. China.}}
\affiliation[2]{organization={Center for Strategic Assessment and Consulting, Academy of Military Science}, 
                 postcode={100091}, 
                 city={Beijing}, 
                 country={P.R. China.}}                 

\begin{abstract}
Momentum methods, such as heavy ball method~(HB) and Nesterov's accelerated gradient method~(NAG), have been widely used in training neural networks by incorporating the history of gradients into the current updating process.
In practice, they often provide improved performance over (stochastic) gradient descent~(GD) with faster convergence.
Despite these empirical successes, theoretical understandings of their accelerated convergence rates are still lacking.
Recently, some attempts have been made by analyzing the trajectories of gradient-based methods in an over-parameterized regime, where the number of the parameters is significantly larger than the number of the training instances.
However, the majority of existing theoretical work is mainly concerned with GD and the established convergence result of NAG is inferior to HB and GD, which fails to explain the practical success of NAG.
In this paper, we take a step towards closing this gap by analyzing NAG in training a randomly initialized over-parameterized two-layer fully connected neural network with ReLU activation.
Despite the fact that the objective function is non-convex and non-smooth, we show that NAG converges to a global minimum at a non-asymptotic linear rate $(1-\Theta(1/\sqrt{\kappa}))^t$, where $\kappa > 1$ is the condition number of a gram matrix and $t$ is the number of the iterations. 
Compared to the convergence rate $(1-\Theta(1/{\kappa}))^t$ of GD, our result provides theoretical guarantees for the acceleration of NAG in neural network training.
Furthermore, our findings suggest that NAG and HB have similar convergence rate.
Finally, we conduct extensive experiments on six benchmark datasets to validate the correctness of our theoretical results.
\end{abstract} 
\begin{keyword}
Neural networks\sep Over-parameterization \sep Neural tangent kernel \sep Nesterov's accelerated gradient method  \sep Non-asymptotic global convergence
\end{keyword}
\maketitle

\section{Introduction}
Momentum methods play a crucial role in numerous areas, including machine learning~\cite{Lin2020}, signal processing~\cite{beck2009fast}, and control~\cite{Qu}.
Typical momentum techniques, including  heavy-ball method~(HB)~\cite{polyak1964some} and Nesterov's accelerated gradient method~(NAG)~\cite{nesterov1983method}, improve the performance of gradient descent~(GD) for tackling convex tasks both in theoretical and empirical performance.
In the case of a quadratic strongly convex problem, HB has an accelerated convergence rate compared to GD~\cite{polyak1964some}, implying that HB requires fewer iterations than GD to reach the same training error.
In 1983, Nesterov~\cite{nesterov1983method} proposed the NAG method and proved that it has the optimal convergence rate for convex problem with Lipschitz gradient. 

Given the success of momentum methods in convex optimization, they have also been widely adopted in training neural networks for faster convergence~\cite{sutskever2013importance,dozat2016incorporating,ma2018quasi}.
Nowadays, many popular modern methods have taken advantage of momentum techniques, such as Adam~\cite{kingma2014adam}, AMSGrad~\cite{reddi2019convergence}, and AdaBound~\cite{luo2019adaptive}.
In many popular deep learning libraries, momentum methods and their variants are implemented as the default optimizers~\cite{DBLP:conf/nips/PaszkeGMLBCKLGA19, gulli2017deep, DBLP:conf/osdi/AbadiBCCDDDGIIK16}.
Nonetheless, the optimization problem for the neural network is both non-convex and non-smooth due to the usage of the non-linear activation functions. 
In general, it is NP-hard to obtain the global-optimal solution for handling non-convex problems~\cite{DBLP:journals/mp/MurtyK87}. 
From a theoretical view, it remains unclear whether momentum methods are capable of learning a neural network with low training loss, let alone the acceleration of momentum methods over GD.

Recently, some theoretical progress has been made towards bridging this gap by analyzing the convergence of (stochastic) GD for training an over-parameterized two-layer ReLU neural network~\cite{du2018gradient,li2018learning,DBLP:conf/icml/DuLL0Z19,DBLP:conf/icml/Allen-ZhuLS19,arora2019fine,song2019quadratic}, where the number of the parameters is much larger than that of the training data. 
The main idea is to investigate the trajectory of gradient-based methods via a kernel matrix called neural tangent kernel~(NTK), which was first introduced by Jacot~\cite{NIPS2018_8076} to study the optimization of infinite wide neural networks.
However, most existing literature is concerned with GD.
To our knowledge, there are only two recent papers on the convergence of momentum methods in training neural networks~\cite{wang2020provable, bu2020dynamical}.
Focusing on a discrete-time setting, Wang \textit{et al.}~\cite{wang2020provable} proved HB is able to achieve a linear convergence rate to the global optimum and attains an acceleration beyond GD.
From a continuous-time perspective, Bu \textit{et al.}~\cite{bu2020dynamical} found a similar result for HB.
Nevertheless, their analysis relies on the approximation between a second-order ordinary differential equation~(ODE) and the momentum method with an infinitesimal learning rate, which is far from practical implementations.
Moreover, their result showed that NAG with a time-varying momentum coefficient converges at an asymoptotic sublinear rate, which is inferior to GD~\cite{du2018gradient,wu2019global} and HB~\cite{wang2020provable}.
In contrast, when optimizing a neural network, it was empirically observed that NAG outperforms GD and exhibits comparable (even better) performance compared to HB~\cite{sutskever2013importance, DBLP:conf/icml/SchmidtSH21}.
Therefore, there is a lack of enough understandings about the acceleration of NAG.

In this work, we consider training a randomly initialized over-parameterized two-layer ReLU neural network with NAG.
In fact, there are several variants of NAG proposed by Nesterov~\cite{nesterov2003introductory}.
We focus on NAG with a constant momentum parameter, which is the default scheme of NAG implemented in PyTorch~\cite{DBLP:conf/nips/PaszkeGMLBCKLGA19}, Keras~\cite{gulli2017deep} and TensorFlow~\cite{DBLP:conf/osdi/AbadiBCCDDDGIIK16}.
Inspired by~\cite{du2018gradient,wang2020provable},  we exploit the connection between the NTK and the wide neural network to establish theoretical convergence guarantees for NAG.
Specifically, our contributions can be summarized as follows:
\begin{enumerate}
	\item Firstly, we intuitively show that the residual dynamics of an infinite width neural network trained by NAG can be approximated by a linear discrete dynamical system, whose coefficient matrix is determined by NAG's hyperparameters and the NTK matrix.
	When the spectral norm of the coefficient matrix is less than 1, NAG is able to attain a global minimum at an asymptotic linear convergence rate according to Gelfand's formula~\cite{1941Normierte}. 
	\item Secondly, borrowing the idea from the infinite width case, we establish the residual dynamics of NAG in training a finite width neural network.
	 By analyzing the dynamics, we show that NAG converges to a global minimum at a non-asymptotic rate $(1-\Theta(1/\sqrt{\kappa}))^t$, where $\kappa > 1$ is the condition number of the NTK matrix and $t$ is the number of the iterations.
	 Moreover, compared to the convergence rate $(1-\Theta(1/{\kappa}))^t$ of GD~\cite{du2018gradient,wu2019global}, our result provides theoretical guarantees for the acceleration of NAG over GD. 
	\item Thirdly, we demonstrate that NAG exhibits a different residual dynamics compared to HB~\cite{wang2020provable}, but the corresponding coefficient matrix shares a similar spectral norm, which results in a comparable convergence rate as HB.
Our analysis of the residual dynamics induced by NAG is of independent interest and may further extend to study other NAG-like algorithms and the convergence of NAG in training other types of neural network.
	\item  Finally, we conduct extensive experiments on six benchmark datasets.
	In the convergence analysis, we empirically show that NAG outperforms GD and obtains a comparable and even better performance compared to HB, which verifies our theoretical results.
	Furthermore, using all six datasets, we investigate the impact of the over-parameterization on two quantities related to our proof. The result also suggests the correctness of our findings. 
\end{enumerate}

\section{Related work}
\textbf{First-order methods.} 
With the growing demands for handling large-scale machine learning problems, 
first-order methods that only access the objective values and gradients have become popular due to their efficiency and effectiveness.

For convex problems, GD is the most well-known first-order method, which achieves $\mathcal{O}(1/t)$ convergence rate with $t$ iterations~\cite{nesterov2003introductory}.
Momentum methods make a further step by exploiting the history of gradients. 
Among first-order methods, NAG obtains the optimal rate $\mathcal{O}(1/t^2)$ for convex problem with Lipschitz gradient~\cite{nesterov2003introductory}.
Focusing on non-smooth convex problems, Tao \textit{et al.}~\cite{DBLP:journals/tnn/TaoPWT20} proved that NAG improves the convergence rate of stochastic gradient descent by a factor $\log(t)$.
In contrast, Lessard \textit{et al.}~\cite{lessard2016analysis} found a counterexample that HB may fail to find the global optimum for some strongly convex problems.
On the other hand, several researches established a connection between the discrete-time methods and the ODE models.
In the limit of infinitesimally learning rate, Su \textit{et al.}~\cite{JMLR:v17:15-084} formulated a second-order ODE associated with NAG.
The convergence of NAG is then linked to the analysis of the related ODE solution.
Shi \textit{et al.}~\cite{DBLP:journals/corr/abs-1810-08907} further developed a more accurate high-resolution ODE that helps distinguish between HB and NAG.

For non-convex problems, it is intractable to find a global optimum.
As an alternative, current researches consider the convergence to the stationary point or local minimum as a criterion for evaluation~\cite{DBLP:journals/mp/CarmonDHS20,DBLP:conf/icml/Jin0NKJ17,DBLP:conf/icml/CarmonDHS17,DBLP:journals/siamjo/DiakonikolasJ21}.
In contrast to previous work, we show a non-asymptotic convergence result for NAG to arrive at a global minimum for a non-convex and non-smooth problem.

\textbf{Convergence theory of over-parameterized neural networks.}
Du \textit{et al.}~\cite{du2018gradient} was the first to prove  the convergence rate of GD for training a randomly initialized two-layer ReLU neural network.
Their results showed that GD can linearly converge to a global optimum when the width of the hidden layer is large enough.
Based on the same neural network architecture, Li and Liang~\cite{li2018learning} investigated the convergence of stochastic gradient descent on structured data.
Wu \textit{et al.}~\cite{wu2019global} improved the upper bound of the learning rate in~\cite{du2018gradient}, which results in a faster convergence rate for GD. 
On the other hand, Jacot~\textit{et al.}~\cite{NIPS2018_8076} introduced the NTK theory, which establishes a link between the over-parameterized neural network and the neural tangent kernel.
Their result was further extended to investigate the convergence of GD for training different architectures of neural networks, including convolutional~\cite{arora2019exact}, residual~\cite{DBLP:journals/corr/abs-2002-06262} and graph neural network~\cite{DBLP:conf/nips/DuHSPWX19}.
While these results are mostly concerned with GD, there are few theoretical guarantees for momentum methods. 

Recently, some researchers have drawn attention to analyzing the convergence of  momentum methods with NTK theory.
Wang \textit{et al.}~\cite{wang2020provable} studied the convergence of  HB using a similar setting as~\cite{du2018gradient}.
They proved that, as compared to GD, HB converges linearly to the global optimum at a faster rate.
Bu \textit{et al.}~\cite{bu2020dynamical} established the convergence results of HB and NAG by considering their limiting ODE from a continuous perspective.
Nonetheless, their analysis is asymptotic and far from practice because they use the infinitesimal learning rate and the approximation of Dirac delta function.
In contrast, our analysis focuses on the discrete-time situation and yields a non-asymptotic convergence rate of NAG with a finite learning rate, which is close to the reality. 

Furthermore, some researchers applied optimal transport theory to analyze  the training dynamics of neural networks in the mean field setting~\cite{DBLP:journals/corr/abs-1804-06561, chizat2018global}, where the evolution of the parameter can be approximated by a distributional dynamics.
However, their results are limited to (stochastic) GD.

\section{Preliminaries}
\subsection{Notation}
In the paper, we use lowercase, lowercase boldface and uppercase boldface letters to represent  scalars, vectors and matrices, respectively.
Let $[n]$ denote $\{1, 2, \cdots, n\}$.
For any set $S$, let $|S|$ be its cardinality.
Let $\|\cdot\|$ be the $\ell_2$ norm of the vector or the spectral norm of the matrix, and $\|\cdot\|_F$ be the Frobenius norm.
We denote $\langle \cdot, \cdot \rangle$ as the Euclidean inner product.
We use $\lambda_{max}(\textbf{X})$, $\lambda_{min}(\textbf{X})$ and $\kappa(\textbf{X})$ to denote the largest eigenvalue, smallest eigenvalue and condition number of the matrix $\textbf{X}$, respectively.
For initialization, we use $\N(0, \textit{I})$ and $Rademacher(1/2)$ to denote the standard Gaussian distribution and the Rademacher distribution, respectively.
We adopt $\mathbb{I}\{\omega\}$ as the indicator function, which outputs 1 when the event $\omega$ is true and 0 otherwise.
The training dataset is denoted by $\mathcal{D} = \{\x_i, y_i\}_{i=1}^n$, where $\x_i \in \R^d$ and $y_i \in \R$ are the features and label of the $i$-th sample, respectively. 
For two sequences $\{a_n\}$ and $\{b_n\}$, we write $a_n = \mathcal{O}(b_n)$ if there exists a positive constant $0 < C_1 < +\infty$ such that $a_n \leq C_1 b_n$, write $a_n = \Omega(b_n)$ if there exists a positive constant $0 \leq C_2 < + \infty$ such that $a_n \geq C_2 b_n$, and write $a_n = \Theta(b_n)$ if there exists two positive constants $0 < C_3, C_4 < +\infty$ such that $a_n \leq C_3 b_n$ and $b_n \leq C_4 a_n$.

\subsection{Problem setting}
\label{problem setting}

In this subsection, we first briefly introduce the update procedures of three commonly used methods: GD, HB and NAG.
Then we provide the details of the architecture and the initialization scheme of the neural network.
Finally, we introduce the main idea of the NTK theory~\cite{NIPS2018_8076}.

\textbf{GD, HB and NAG.} In this paper, we mainly focus on the supervised learning problem in the deterministic setting.
Our aim is to train a model $f: \R^d \to \R$ to predict unobserved features correctly.
The parameter of the model denotes by $\thetab$.
In order to estimate $\thetab$, the common approach is to solve the objective function $\L$ defined on the training dataset $\mathcal{D}$ as:
\begin{eqnarray}
\label{empirical risk}
	\mathop{\min}_{\thetab} \L(\thetab) = \frac{1}{n}\sum_{i=1}^n \ell(y_i, f(\thetab; \x_i)),
\end{eqnarray}
where $\ell: \R \times \R \to \R$ denotes the loss function.
The above problem is also referred to as empirical risk minimization~(ERM). 
Meanwhile, current machine learning problems often involve large-scale training datasets and complex models.
GD has become a common choice due to its simplicity and efficiency, which updates the parameter $\thetab$ as
\begin{eqnarray}
	\thetab_{t+1} = \thetab_t - \eta \nabla \L(\thetab_t),
\end{eqnarray}
where $\eta > 0$ is the learning rate and $\nabla \L(\thetab_t)$ is the gradient with respect to the parameter at the $t$-th iteration.

HB starts from the initial parameter $\thetab_{-1} = \thetab_0$ and updates as follows:
\begin{equation}
	\label{eq:HB_update}
\thetab_{t+1} = \thetab_t + \beta(\thetab_t - \thetab_{t-1}) - \eta \nabla \L(\thetab_t),	
\end{equation}
where $\beta \in [0, 1)$ is the momentum parameter.
NAG is another important development of momentum methods and has several types~\cite{nesterov2003introductory}.
In this paper, we focus on NAG with a constant momentum parameter $\beta$.
Given the initial parameters $\thetab_0$ and  $\v_0 = \thetab_0$,
NAG involves the update procedures in two steps
\begin{eqnarray}
	\label{eq:NAG-SC}
		{\v}_{t+1} &=&  \thetab_t - \eta \nabla \L(\thetab_t) \\
		\thetab_{t+1} &=& {\v}_{t+1} + \beta({\v}_{t+1}-{\v}_{t}),
\end{eqnarray}
which can be reformulated in a equivalent form without $\v$
\begin{eqnarray}
	\label{eq:NAG-SC_one_line}
	\thetab_{t+1} &=& \thetab_t  + \beta(\thetab_t - \thetab_{t-1}) - \eta \nabla \L(\thetab_t) - \beta\eta(\nabla \L(\thetab_t) - \nabla \L(\thetab_{t-1})).
\end{eqnarray}
Compared with HB~(\ref{eq:HB_update}), NAG has an additional term $\beta\eta(\nabla \L(\thetab_t) - \nabla \L(\thetab_{t-1}))$, which computes the difference between two consecutive gradients and is referred to as gradient correction~\cite{shi2019acceleration}.
\begin{figure}[!t]
	\label{architecture}
	\centering
	\includegraphics[scale=0.3]{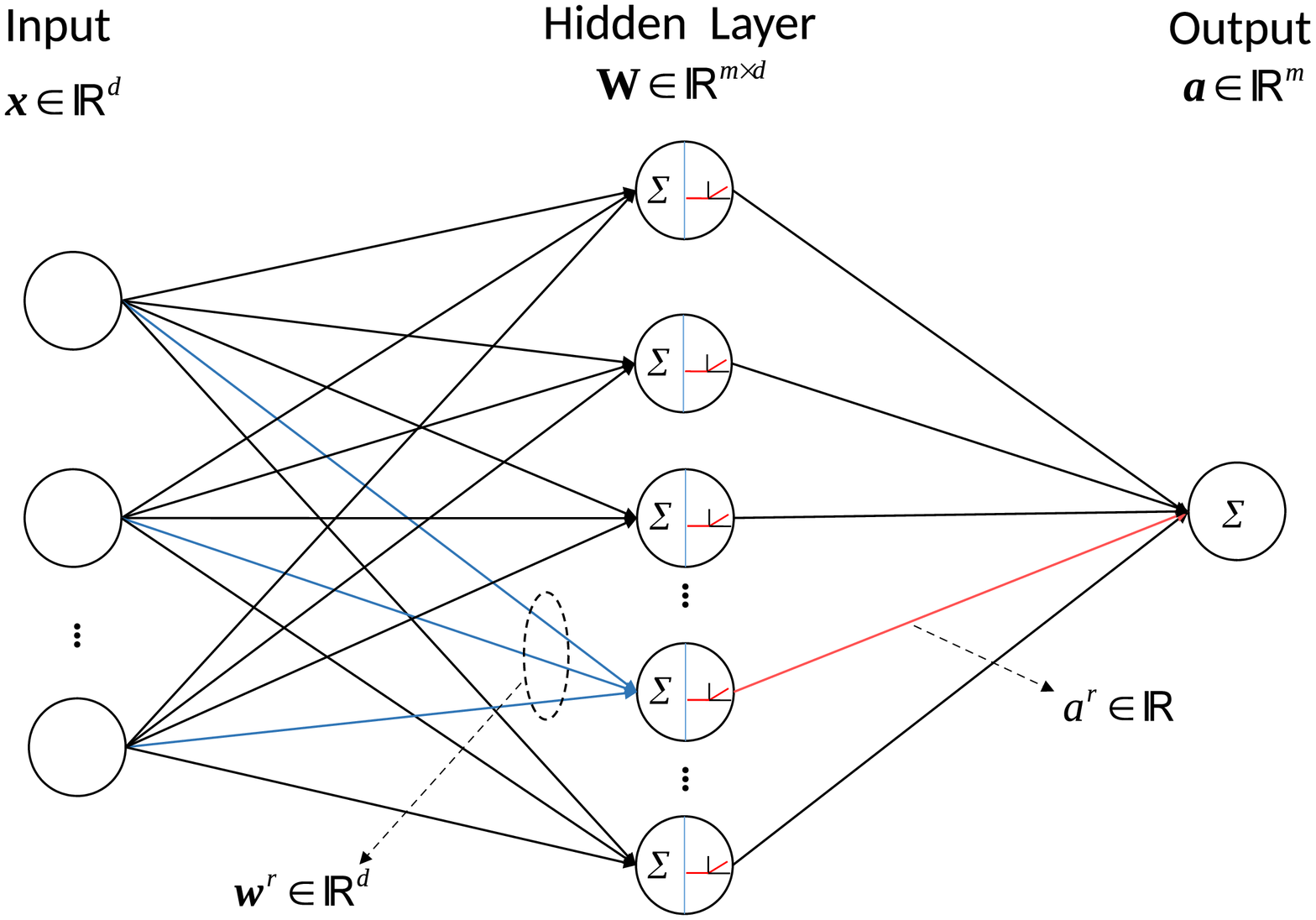}       
	\caption{The architecture of the two-layer fully connected neural network with ReLU activation.}
	\label{interval}
\end{figure}

\textbf{The details of the neural network.} In this work, we consider a two-layer fully connected neural network $f: \R^d \to \R$ as follows:
\begin{equation}
	\label{eq:two_layer neural network}
	f(\W, \textbf{a};\x)= \frac{1}{\sqrt{m}} \sum_{r=1}^m a^r \sigma(\langle \w^r, \x \rangle),
\end{equation}
where $\x \in \R^d$ denotes the input features, $\W=(\w^1, \w^2, \cdots,\w^m) \in \R^{d \times m}$ denotes the weight matrix of the hidden layer, $\textbf{a} = (a^1,a^2,\cdots, a^m) \in \R^m $ denotes the output weight vector and $\sigma(z)=z \cdot \mathbb{I}\{z \geq 0\}$ denotes the ReLU activation function.	
Figure~1 shows the architecture of the neural network. 
The parameters follow the random initialization scheme as $\w^r \sim \N(0, \textit{I}_d)$ and $a^r \sim Rademacher(1/2)$ for any $r \in [m]$.

Following the settings in~\cite{du2018gradient,wang2020provable,arora2019fine}, we keep the output layer $\a$ fixed after initialization and only optimize the weight matrix $\W$ through minimizing the square loss
\begin{equation}
\label{eq:objective}
\L(\W,\a) = \frac{1}{2}\sum_{i=1}^n (y_i - f(\W, \textbf{a};\x_i))^2.
\end{equation} 
Then the gradient for the weight vector of the $r$-th neuron can be calculated as:
\begin{equation}
	\label{eq:gradient_objective}
	\frac{\partial \L(\W,\a)}{\partial \w^r} = \frac{1}{\sqrt{m}}\sum_{i=1}^n (f(\W, \textbf{a};\x_i) - y_i) a^r \x_i \mathbb{I}\{{\langle	\w^r, \x_i \rangle \geq 0}\}.
\end{equation}

Although this model is a simple two-layer neural network, its loss landscape is still non-convex and non-smooth due to the use of ReLU activation function.
However, the objective function $\L$ becomes convex when the weight matrix $\W$ is fixed and just optimizes the output layer $\a$, and this setting has been studied in~\cite{ICML-2018-Nguyen0}.

\begin{algorithm}[!t]\caption{Training Two-Layer Fully Connected ReLU Neural Network with NAG.}\label{alg:alg_main_text}
\begin{algorithmic}[1]
\State \textbf{Parameters}: learning rate $\eta > 0$, momentum parameter $0\leq\beta<1$.
\State \textbf{Initialization}: $\v^r(0) = \w^r(0) \sim \N(0,I_d)$, $a^r \sim Rademacher(1/2)$ for $r\in [m]$. 
\For{$t = 0, \ldots, T$}
	\For{$ r=1, \ldots, m$}
        \State Calculate gradient $\frac{\partial \L(\w(t), \a)}{\partial \w^r(t)}$ for  $\w^r$ using~(\ref{eq:gradient_objective}).
        \State Update $\v^r$: $\v^r(t+1) = \w^r(t) - \eta \frac{\partial \L(\w(t),\a)}{\partial \w^r(t)}$.
        \State Update $\w^r$: {{$\w^r(t+1) = \v^r(t+1) + \beta (\v^r(t+1) - \v^r(t))$}}.
	\EndFor
\EndFor
\end{algorithmic}
\end{algorithm}
\textbf{NTK theory.}  This theory was first introduced by Jacot~\cite{NIPS2018_8076} to study the optimization of infinite wide neural networks.
It is closely related to a Gram matrix $\H_t$, which is defined as: 
\begin{equation}
 \label{eq:NTK}
 \H_t(\x_i, \x_j)=\langle \nabla_{\thetab} f(\thetab_t;\x_i) , \nabla_{\thetab} f(\thetab_t;\x_j) \rangle, \forall \; (i,j) \in [n]\times [n],
\end{equation}
where $f$ is the neural network model and $\theta$ represents its parameter.
Clearly, $\H_t$ is positive semi-definite due to the property of Gram matrix and varies according to $\theta_t$.
As the width of the neural network goes to infinity, 
the limit matrix $\bar{\H} := \lim_{m\to\infty}\H_0$ is 
determined by the initialization and architecture of the corresponding neural network $f$,
which is the so-called NTK matrix.
When the neural network is sufficiently over-parameterized, $\thetab_t$ barely changes from its initial $\thetab_0$, which in turn guarantees $\H_t$ stays close to $\bar{\H}$ during training~\cite{du2018gradient,NIPS2018_8076,arora2019fine}.
As a result, the over-parameterized neural network behaves similarly to its linearization around $\thetab_0$.

Given the specific two-layer neural network~(\ref{eq:two_layer neural network}) and the objective function~(\ref{eq:objective}), it has the corresponding $\H_t$ as:
\begin{eqnarray}
\label{eq:gram matrix h_t}
	\H_t(\x_i, \x_j)=\frac{1}{m}\sum_{r=1}^m \langle \x_i, \x_j \rangle \mathbb{I}\{\langle \w_t^r, \x_i \rangle \geq 0 \& \langle \w_t^r, \x_j \rangle \geq 0\},
\end{eqnarray}
and the NTK $\bar{\H}$ can be calculated with the expected value
\begin{eqnarray}
\label{limiting NTK}
	\bar{\H}(\x_i, \x_j) &=& \E_{\w \sim N(0, \textit{I})}[\langle \x_i, \x_j \rangle \mathbb{I}\{\langle \w, \x_i \rangle \geq 0 \& \langle \w, \x_j \rangle \geq 0\}] \nonumber\\ 
	&=& \langle \x_i, \x_j \rangle \frac{\pi - arccos(\langle \x_i, \x_j \rangle)}{2\pi}.
\end{eqnarray}
In addition, the above $\bar{\H}$ is strictly positive when the training dataset satisfies $\x_i \neq \x_j$ for all $i \neq j$~\cite{du2018gradient}.

\section{Main results}

In this section, we analyze the dynamics of NAG's residual error from a discrete view and give a non-asymptotic convergence rate with specific learning rate and momentum parameter, which is inspired by~\cite{du2018gradient} and \cite{wang2020provable}.

\subsection{Intuition behind our proof}
\label{section4_1}
To start with, we intuitively illustrate the main idea of our proof  under the infinite width assumption.
As mentioned in Section~\ref{problem setting}, some theoretical and empirical works~\cite{NIPS2019_8559,NIPS2019_9063} have shown that the outputs of the over-parameterized neural network can be approximated by its first-order Taylor expansion around its initial parameter as:
\begin{equation}
	\label{eq:linearzation}
	f(\thetab;\x) \approx f(\thetab_0;\x) + \langle\nabla f(\thetab_0;\x),\thetab - \thetab_0 \rangle.
\end{equation}
By taking the derivative on both sides of~(\ref{eq:linearzation}), it has
\begin{eqnarray}
\label{eq:approx gradient}
	\nabla_{\thetab} f(\thetab;\x) \approx \nabla_{\thetab} f(\thetab_0;\x).
\end{eqnarray}
For simplicity, let $\X=(\x_1, \cdots, \x_n) \in \R^{d \times n}$ and $\Y= (y_1, \cdots, y_n) \in \R^n$ be the concatenation of the features and the corresponding labels of dataset $\mathcal{D}$. 
In addition, we define $f(\thetab; \X) = (f(\thetab; \x_1), \cdots, f(\thetab; \x_n)) \in \R^n$,$\nabla_{\thetab} f(\thetab; \X) = (\nabla_{\thetab}f(\thetab; \x_1), \cdots, \nabla_{\thetab}f(\thetab; \x_n))^{\top} \in \R^{n \times  k}$ and $\xib =(f(\thetab; \x_1) -y_1, \cdots, f(\thetab; \x_n) - y_n) \in \R^n$  as the concatenated  outputs, gradients and residual errors of the neural network, respectively.
Plugging NAG's update rule~(\ref{eq:NAG-SC_one_line}) into (\ref{eq:linearzation}), it has
\begin{eqnarray}
	\label{eq:NAG_transform}
	&&f(\thetab_{t+1};\X)\nonumber \\
	\!\!\!\!\!\!&\approx& \!\!\!\! f(\thetab_0;\X) \!+\! \nabla_{\thetab} f(\thetab_0;\X)\big(\thetab_t  \!-\! \eta \nabla_{\thetab} \L(\thetab_t) \!+\! \beta(\thetab_t \!-\! \thetab_{t\!-\!1}) \!-\! \eta\beta\big(\nabla_{\thetab} \L(\thetab_t) \!-\! \nabla_{\thetab} \L(\thetab_{t\!-\!1}) \big)  \!-\! \thetab_0 \big) \nonumber\\
	\!\!\!\!\!\!&\approx&\!\!\!\! f(\thetab_t;\X)  - \eta\beta\nabla_{\thetab} f(\thetab_0;\X)\big(\nabla_{\thetab} \L(\thetab_t) -\nabla_{\thetab} \L(\thetab_{t-1}) \big) \!-\! \eta \nabla_{\thetab} f(\thetab_0;\X)\nabla_{\thetab}\L(\!\thetab_t;\X) \nonumber\\
	 &+& \!\!\!\! \beta \big( f(\thetab_t;\X) \!-\!f(\thetab_{t-1};\X) \big)	  , 
\end{eqnarray}
where the last approximation uses~(\ref{eq:linearzation}).
Expanding $\nabla_{\thetab}\L(\thetab_t)$ with~(\ref{eq:objective}), it has
\begin{eqnarray}
\label{eq:expansion of L}
	\nabla_{\thetab}\L(\thetab_t) = \nabla_{\thetab}f(\thetab_t;\X)^{\top}(f(\thetab_t; \X) - \Y).
\end{eqnarray}

Then, plugging (\ref{eq:approx gradient}) and (\ref{eq:expansion of L}) into (\ref{eq:NAG_transform}), it has the approximated residual error as:
\begin{eqnarray}
	\xib_{t+1} \!\!\!&=&\!\!\! f(\thetab_{t+1};\X) - \Y \nonumber \\
	\label{eq:NAG_approx_one}
	\!\!\!&\approx&\!\!\!\!\! \xib_t \!-\! \eta \H_0\xib_t \!+\! \beta(\xib_t \!-\! \xib_{t-1}) \!-\!\eta\beta\H_0(\xib_t-\xib_{t-1}). 
\end{eqnarray}
Reformulating~(\ref{eq:NAG_approx_one}), it has
\begin{eqnarray}
\label{eq:residual_recurisvie}
 \begin{bmatrix} \xib_{t+1} \\ \xib_t \end{bmatrix} \approx \begin{bmatrix}
   (1\!+\!\beta)(\I_n\!-\!\eta \H_0) \!&\!
  \beta(\!-\I_n\!+\!\eta \H_{0}) \\
   \I_n \!&\! \textbf{0}_n 
   \end{bmatrix}	\begin{bmatrix} \xib_{t} \\ \xib_{t-1} \end{bmatrix},
\end{eqnarray}
where $\I_n \in \R^{n\times n}$ and $\textbf{0}_n \in \R^{n \times n}$ denote the identity matrix and zero matrix, respectively.
Similar to the quadratic convex optimization case~\cite{lessard2016analysis,flammarion2015averaging}, the residual error in~(\ref{eq:residual_recurisvie}) follows a linear dynamical system.
When the spectral norm of the coefficient matrix in~(\ref{eq:residual_recurisvie}) is less than one, the residual error decays to zero at an asymptotic linear convergence rate according to Gelfand's formula~\cite{1941Normierte}.
However, this result is asymptotic and depends on the infinite width assumption.
In contrast, we rely on a mild assumption about the width and provide a non-asymptotic convergence result.

\subsection{Residual dynamics for NAG}
Our analysis depends on an important event: $$A_{ir} = \{\exists \w: \|\w - \w_0^r\| \leq R, \mathbb{I}\{\langle \w, \x_i \rangle\} \neq \mathbb{I}\{\langle \w_0^r, \x_i \rangle\}\},$$
where $R>0$ is a constant.
This event describes whether there exists a weight vector that restricts in a neighbourhood of its initial  but has a different activation pattern compared to the initial for the same input.
Here, the activation pattern is defined as the output of $\mathbb{I}\{\langle \w, \x \rangle\}$.
Then one can define the set $S_i = \{r\in [m]: \mathbb{I}\{A_{ir}\} = 0\}$ and its complementary set $S_i^{\perp} = [m] \setminus S_i $ to separate the neurons with two parts.

By utilizing the intuition introduced in Section~\ref{section4_1},  we provide the recursion formulation of the residual error for a finite width two-layer neural network trained by NAG.

\begin{lemma}
\label{lemma: rec form}
Let $\xib_t$ be the residual error vector of the $t$-th iterate in NAG for any $t \in [T]$, it has 
\begin{equation}
\label{eq:recursion}
	{{\xib_{t+1} = \xib_t - \eta \H_0\xib_t + \beta(\xib_t - \xib_{t-1}) -\eta\beta\H_0(\xib_t -\xib_{t-1}) +\psib_t + \phib_t}}, 
\end{equation}
where 
\begin{eqnarray}
\label{eq: the definition of psib}
	\psib_t= \beta\eta(\H_{t-1} - \H_0)\xib_{t-1} -(1+\beta)\eta(\H_t - \H_0)\xib_t,
\end{eqnarray}
and  the $i$-th element of $\phib_t$ is bounded by
\begin{equation}
\label{eq:bound of phib}
	|\phib_t[i]| \leq \frac{ \sup_{j\in [n]}|S_j^{\perp}|\sqrt{n}\eta}{m}\left[(2+4\beta)\|\xib_t\|+ 3\beta\|\xib_{t-1} \|+2\sum_{i=0}^{t-1}\beta^{t+1-i}\|\xib_i\|\right]. 
\end{equation}
\end{lemma}
The proof of Lemma 1 can be found in~\ref{app:lemma_1}.
Denotes $\z_t = [\xib_t; \xib_{t-1}]$ as the augmented residual error  at iteration $t$,
then~(\ref{eq:recursion}) can be reformulated as:
\begin{eqnarray}
\label{eq:matrix_form_residual}
	\z_{t+1} = \M \z_t + \mub_t,
\end{eqnarray}
where $\mub_t = [\psib_t+ \phib_t;\textbf{0}]$ and the coefficient matrix
$\M=\begin{bmatrix}
   (1\!+\!\beta)(\I_n\!-\!\eta \H_0) \!&\!
  \beta(\!-\I_n\!+\!\eta \H_0) \\
   \I_n \!&\! \textbf{0}_n 
   \end{bmatrix}$.
   
Note that, compared to the linear dynamical system~(\ref{eq:residual_recurisvie}), the finite width one has an additional term $\mub$, which can be regarded as a perturbation and we will discuss its bound later.
Furthermore, as shown in~\cite{wang2020provable}, HB has the residual dynamics as
{\small{\begin{eqnarray}
\label{eq::matrix_form_HB}
\begin{bmatrix} \xib_{t+1} \\ \xib_t \end{bmatrix} = \begin{bmatrix}
   (1\!+\!\beta)\I_n\!-\!\eta \H_0 \!&\!
  -\beta\H_{0}) \\
   \I_n \!&\! \textbf{0}_n 
   \end{bmatrix}	\begin{bmatrix} \xib_{t} \\ \xib_{t-1}\end{bmatrix} + \mub_t^{'},
\end{eqnarray}}}
which differs from~(\ref{eq:matrix_form_residual}) both in the  coefficient matrix and perturbation term.

\subsection{Convergence analysis}
\label{main_theory_convergence}
\renewcommand\arraystretch{1.3}
\begin{table*}
\caption{Summary of the Convergence Results. 
Let $m$ denotes the width of neural network. 
Let $n$ denotes the number of input data points. 
Let $\delta$ denotes the failure probability. 
Let $t$ denotes the iteration number.
Define $\lambda = \lambda_{min}(\bar{\H})$ , $\lambda_{max} = \lambda_{max}(\bar{\H}) + \lambda/4$ and $\kappa = 4\kappa(\bar{\H})/3 + 1/3$. }
\label{table1}
\centering
\begin{tabular}{ | M{1.3cm}| M{3.3cm}| M{4cm}| M{2.7cm}|} 
\hline
{\bf Method} & Width of the Neural Network & Hyperparameter Choice & Convergence Rate \\ \hline
GD~\cite{wu2019global} & $\Omega(\lambda^{-4} \delta^{-3}n^6)$ & $\eta = \Theta(\frac{1}{\lambda_{max}(\bar{\H})})$ & $(1-\Theta(\frac{1}{\kappa}))^t$  \\ [1.4ex] \hline
HB~\cite{wang2020provable} & $\Omega(\lambda^{-2}n^{4}\kappa^2 \log^3(n/\delta))$ & $\eta = \frac{1}{\lambda_{max}},\beta = (1-\frac{1}{2\sqrt{\kappa}})^2 $ & $(1-\frac{1}{4\sqrt{\kappa}})^t$ \\ [1.4ex] \hline
{NAG} & $\Omega(\lambda^{-2}n^{4}\kappa^2 \log^3(n/\delta))$ & $\eta = \frac{1}{2 \lambda_{max}},\beta = \frac{3\sqrt{\kappa} - 2}{3\sqrt{\kappa} + 2} $ & $(1-\frac{1}{2\sqrt{\kappa}})^t$ \\ [1.4ex] \hline
\end{tabular}
\end{table*}

By recursively using~(\ref{eq:matrix_form_residual}), it has
\begin{eqnarray}
	\label{eq: evolution of z_t}
	\z_t = \M^t \z_0 + \sum_{i=0}^{t-1}\M^{t-1-i} \mub_i.
\end{eqnarray}
Then applying Cauchy-Schwarz inequality on~(\ref{eq: evolution of z_t}), we have
\begin{eqnarray}
\label{eq:matrix_form}
	\|\z_t\| \leq \|\M^t \z_0\| + \|\sum_{i=0}^{t-1}\M^{t-1-i} \mub_i\|.
\end{eqnarray}
In order to prove the convergence of NAG, it needs to separately derive  bounds for the two terms on the right-hand side of (\ref{eq:matrix_form}).
The first term is the norm of the product between the matrix power $\M^t$ and the vector $\z_0$.
We provide its upper bound in the following lemma.

\begin{lemma}
\label{lemma: matrix_vector}
  Assume $\H \in \R^{n \times n}$ is a symmetry positive definite matrix.
Let{\small{ $\M = \begin{bmatrix}
   (1\!+\!\beta)(\I_n\!-\!\eta \H) &
  \beta(-\I_n\!+\!\eta \H) \\
   \I_n & \textbf{0}_n 
   \end{bmatrix} \in \R^{2n \times 2n}$}}.
   Suppose a sequence of iterates $\{\v_i\}$ satisfy $\v_t = \M\v_{t-1}$ for any $t \leq T$.
   If $\beta$ and $\eta$ are chosen that satisfy $1 > \beta \geq \frac{1-\sqrt{\eta\lambda_{min}(\H)}}{1+\sqrt{\eta\lambda_{min}(\H)}}$ and $0 < \eta \leq 1/\lambda_{max}(\H)$, then it has the bound at any iteration $k \leq T$ as
\begin{equation}
\label{eq:the bound of matrix vector}
\|\v_k\| \leq C \big(\sqrt{\beta(1-\eta\lambda_{min}(\H))}\big)^k  \|\v_0\|, 
\end{equation}
where $ C  = \frac{2\beta(1-\eta\lambda_{min}(\H)) + 2}{\sqrt{\min\{g(\beta, \eta\lambda_{min}(\H)), g(\beta, \eta\lambda_{max}(\H))\}}}$ and the function $g$ is defined as $g(x, y) = 4x(1-y) - [(1+x)(1-y)]^2$.
\end{lemma}
The proof is provided in the \ref{app: matrix_vector}.
Given the ranges of the hyperparameters $\eta$ and $\beta$, it is easy to observe that $\sqrt{\beta(1-\eta\lambda_{min}(H))} < 1$, which ensures the decline of  $\|\v_k\|$ during evolution.
For further determining the upper bounds for C and the decay rate, we set $\eta$ and $\beta$ with the spectrum of $\H$.

\begin{lemma}
\label{lemma: specific setting}
Assume $0 < \lambda \leq \lambda_{min}(\H) \leq \lambda_{max}(\H) \leq \lambda_{max}$.
Denote ${\kappa} = \lambda_{max}/\lambda$.
With $\eta = 1/2\lambda_{max}$ and $\beta = \frac{3\sqrt{{\kappa}} - 2}{3\sqrt{{\kappa}} + 2}$, it has
\begin{eqnarray}
\sqrt{\beta(1-\eta\lambda_{min}(\H))} \leq 1 - \frac{2}{3\sqrt{{\kappa}}} ,\;\; C \leq 12\sqrt{{\kappa}}.
\end{eqnarray} 
\end{lemma}

Furthermore, it should be noted that $\M$ in (\ref{eq:matrix_form}) is composed by $\H_0$, which depends on the random initialization of $\W_0$.
For an over-parameterized neural network, the eigenvalues of the random matrix $\H_0$ can be bounded by the spectrum of the deterministic NTK matrix $\bar{\H}$~\cite{wang2020provable}, thereby allowing us to determine the hyperparameters with specific values.
\begin{lemma}(Lemma 13 in \cite{wang2020provable})
\label{lemma: bound of H_0}
Denote $\lambda = \lambda_{min}(\bar{\H})$. Set $m=\Omega(\lambda^{-2}n^2\log(n/\delta))$.
Assume $\w_0^r \sim \N(0, I_d)$ for all $r \in [n]$.
With probability at least $1-\delta$, it holds that
\begin{eqnarray}
\|\H_0 -\bar{\H}\|_F \leq \frac{\lambda}{4}&,&
\lambda_{min}(\H_0) \geq \frac{3}{4}\lambda > 0 \;\; 
 \;\;\lambda_{max}(\H_0) \leq \lambda_{max}(\bar{\H}) + \frac{\lambda}{4}. \nonumber
\end{eqnarray}
As a result, the condition number of $\H_0$ is bounded by
\begin{eqnarray}
 \kappa(\H_0) \leq \frac{4}{3}\kappa(\bar{\H}) + \frac{1}{3}.
\end{eqnarray}
\end{lemma}

Now we turn to analyzing the second term on the right-hand side of~(\ref{eq:matrix_form}), which is also composed by the product of the matrix power $\M^i$ and a bounded vector.
Then it only needs to bound the norm of $\mub$.
Using the Cauchy-Schwarz inequality, it has $\|\mub_t\| \leq \|\phib_t\| + \|\psib_t\|$.

From (\ref{eq:bound of phib}), we observe that the bound of $|\phib_t[i]|$ mainly depends on the term $|S_i^{\perp}|$, which describes how many neurons change their activation patterns on the $i$-th instance during training.
According to recent studies~\cite{du2018gradient,song2019quadratic},
$|S_i^{\perp}|$ has an upper bound $4mR$, which is determined by the distance $R$ between  $\w^r_t$ and its initialization for any $r \in [m]$ and $t \in [T]$.
In~\ref{section: supporting lemmas}, Lemma~\ref{lemma: bound of S_i} presents the details.
When $R$ is small enough, it has $|S_i^{\perp}| \ll m$.

On the other hand, the bound of $\|\psib_t\|$ is closely related to  the distance between $\H_t$ and $\H_0$.
Previous works~\cite{du2018gradient,song2019quadratic} showed that the upper bound of $\|\H_t - \H_0\|$ is also determined by the distance $R$,  
 where Lemma~\ref{lemma: H_t and H_0} in~\ref{section: supporting lemmas} gives the details.

In Theorem 1, we derive $R=\mathcal{O}(1/\sqrt{m})$, which helps control the size of $\|\phib_t\|$ and $\|\psib_t\|$ with an appropriate $m$.
The corresponding bounds for $R$, $\|\phib\|$ and $\|\psib\|$ are given in the proof of Theorem 1.
Finally, we introduce our main result on the convergence of NAG.
\begin{thm}
Define $\lambda = \frac{3\lambda_{min}(\bar{\H})}{4}$, $\lambda_{max} = \lambda_{max}(\bar{\H}) + \frac{\lambda}{4}$ and ${\kappa} = \frac{4}{3}\kappa(\bar{\H}) + \frac{1}{3}$. 
Assume $\w_0^{r} \sim N(0, I_d)$ and $a^r \sim Rademacher(1/2)$ for all $r \in [m]$.
Suppose the number of the nodes in the hidden layer is $m=\Omega(\lambda^{-2}n^{4} \kappa^2 log^3(n/\delta))$.
If the leaning rate $\eta = 1/(2\lambda_{max})$ and the momentum parameter $\beta = \frac{3\sqrt{{\kappa}} - 2}{3\sqrt{{\kappa}} + 2}$, 
with probability at least $1-\delta$ over the random initialization, the residual error for NAG  at any iteration $t$ satisfies
{{\begin{equation}
	\label{eq:theorem_NAG}
\left\|\begin{bmatrix} \xib_t \\ \xib_{t-1} \end{bmatrix}\right\| \leq {(1 - \frac{1}{2\sqrt{{\kappa}}})^t} 2\gamma \textstyle
\left\|\begin{bmatrix} \xib_0 \\ \xib_{-1} \end{bmatrix}\right\| ,
\end{equation}}}
where $\gamma =12\sqrt{{\kappa}}$.

For every $r \in [m]$, we have
\begin{equation}
\label{w_r_distance}
	\|\w_t^r - \w_0^r\| \leq \frac{48\sqrt{2n\kappa}}{\lambda\sqrt{m}}\|\xib_0\| .
\end{equation}
\end{thm}

\textbf{Remark 1}. With the initialization $\W_{-1} = \W_0$, it has $\xib_{-1}=\xib_0$.
Thus, according to Theorem 1,  the training error $\xib_t$ of NAG converges linearly to zero at a $(1-\frac{1}{2\sqrt{\kappa}})^t$ rate after $t$ iteration, which indicates NAG is able to achieve the global minimum as GD and HB.

\textbf{Remark 2}. As shown in~\cite{du2018gradient},  GD converges at a rate $(1-{\frac{\eta\lambda}{2}})^t$, but with a small learning rate $\eta = \mathcal{O}(\frac{\lambda}{n^2})$.
\cite{wu2019global} further improved the bound of the learning rate to $\mathcal{O}(\frac{1}{\|{\bar{\H}}\|})$, where $\|{\bar{\H}}\| \leq n$ and provides an $\mathcal{O}(\lambda/n)$ improvement.
This results in a faster convergence rate $(1-\Theta(1/\kappa))^t$ for GD.
As shown in Theorem 1, NAG obtains a smaller convergence rate $(1 - \Theta(1/\sqrt{\kappa}))^t$, which validates its acceleration over GD.
Moreover, compard to the convergence rate of HB as proved in~\cite{wang2020provable}, 
our results show that NAG obtains a comparable convergence rate.

\textbf{Remark 3}.  The initial residual error satisfies $\|\xib_0\|^2 = \mathcal{O}(nlog(m/\delta)log^2(n/\delta))$ as shown in Lemma~\ref{lemma: init bound}.
Therefore, the upper bound $R$ of $\|\w_t^r - \w_0^r\|$ scales as $\mathcal{O}(1/\sqrt{m})$ for any $r\in[m]$ according to (\ref{w_r_distance}).
This is consistent with the NTK regime that the parameter is hardly changed when the neural network is over-parameterized.
Moreover, the number of the changed activation patterns is bounded by  $|S_i^{\perp}|=|\sum_{r=1}^m \mathbb{I}\{\langle \w_t^r, \x_i \rangle\} \neq \mathbb{I}\{\langle \w_0^r, \x_i \rangle\}| \leq 4mR$ according to Lemma~\ref{lemma: bound of S_i}.
As a result, $\sum_{i \in [n]}|S_i^{\perp}|/(mn)$ can be upper bounded with $4R$, which also scales with $\mathcal{O}(1/\sqrt{m})$.
On the other hand, GD has $R = \mathcal{O}(\frac{\sqrt{n}}{\lambda \sqrt{m}}\|\xib_0\|)$ according to~\cite{du2018gradient}, which is smaller than NAG due to $\kappa > 1$.
Furthermore, $R$ of HB scales as $\mathcal{O}(\frac{\sqrt{n\kappa}}{\lambda \sqrt{m}}\|\xib_0\|)$~\cite{wang2020provable}, which is similar as NAG.

\section{Numerical experiments}
In this section, we conduct extensive experiments to validate our theoretical results, including i) the convergence comparison  between NAG, HB and GD. ii) the impact of the over-parameterization on some quantities as introduced in Remark 3.

\subsection{Setup}
Six benchmark datasets are used in the experiments:   FMNIST~\cite{xiao2017fashion}, MNIST~\cite{lecun1998gradient},  CIFAR10~\cite{krizhevsky2009learning} and three UCI regression datasets (ENERGY, HOUSING and YACHT)~\cite{DBLP:conf/nips/LeeSPAXNS20}. 
The pre-processing of the first three image classification datasets follows the procedures outlined in~\cite{arora2019fine}, where we use the first two classes of images with 10,000 training instances, where the label of the first class is set to +1 and -1 otherwise. 
For all six datasets, we normalize all instances with the unit norm.

According to Table~\ref{table1}, the eigenvalues of the NTK matrix $\bar{\H}$ are used to determine the hyperparameters of each optimizer.
Note that the matrix $\bar{\H}$ is analytic, so its eigenvalues can be easily calculated based on~(\ref{limiting NTK}).
As described in Section~\ref{problem setting}, we use the same architecture and the initialization scheme of the neural network, which is trained with the square loss (\ref{eq:objective}) in the deterministic setting. 
All experiments are conducted on 8 NVIDIA Tesla A100 GPU and the code is written in JAX~\cite{jax2018github}.

\begin{figure*}[!t]
	\centering
	\subfigure[FMNIST]{\includegraphics[scale=0.47]{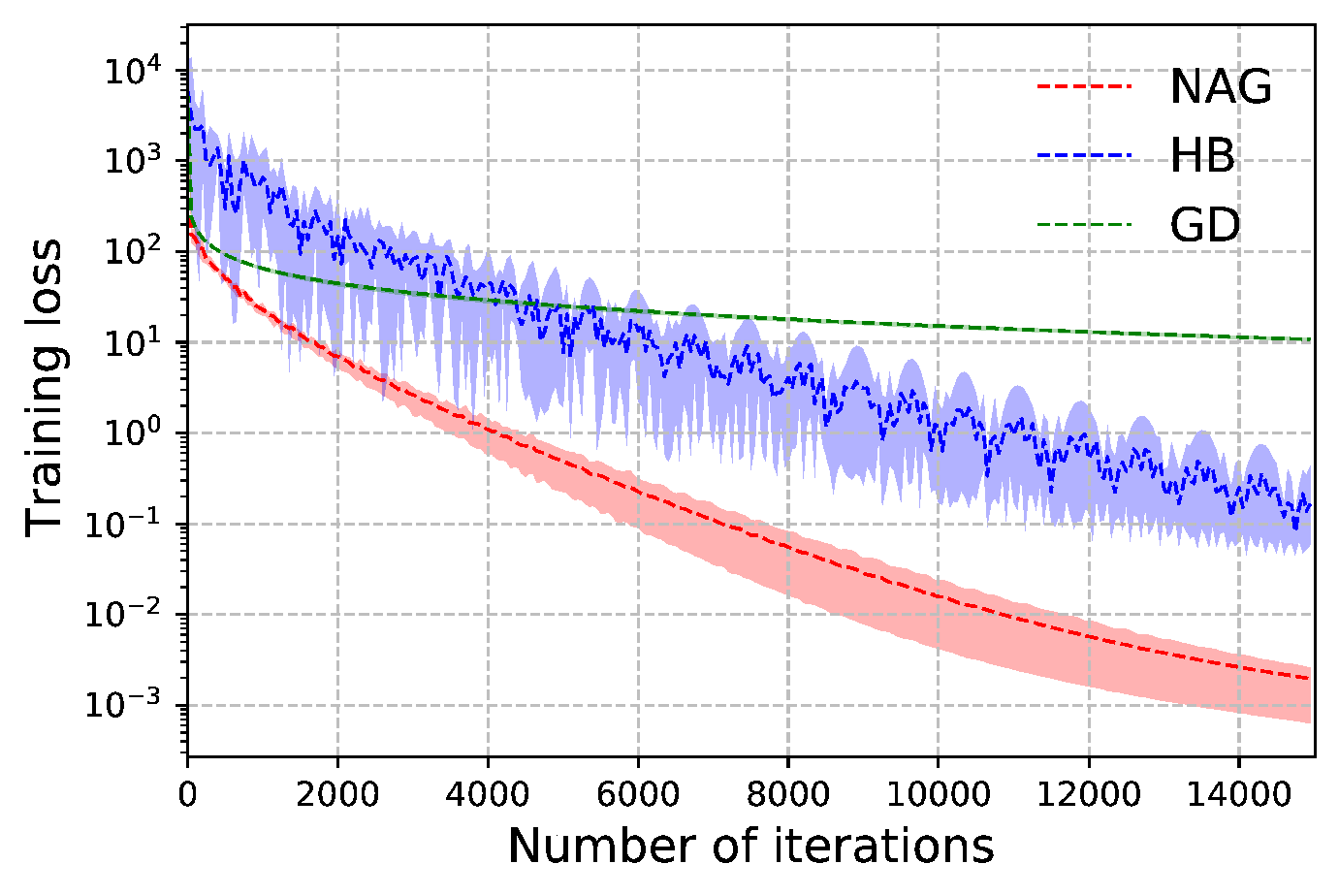}}
	\subfigure[MNIST]{\includegraphics[scale=0.47]{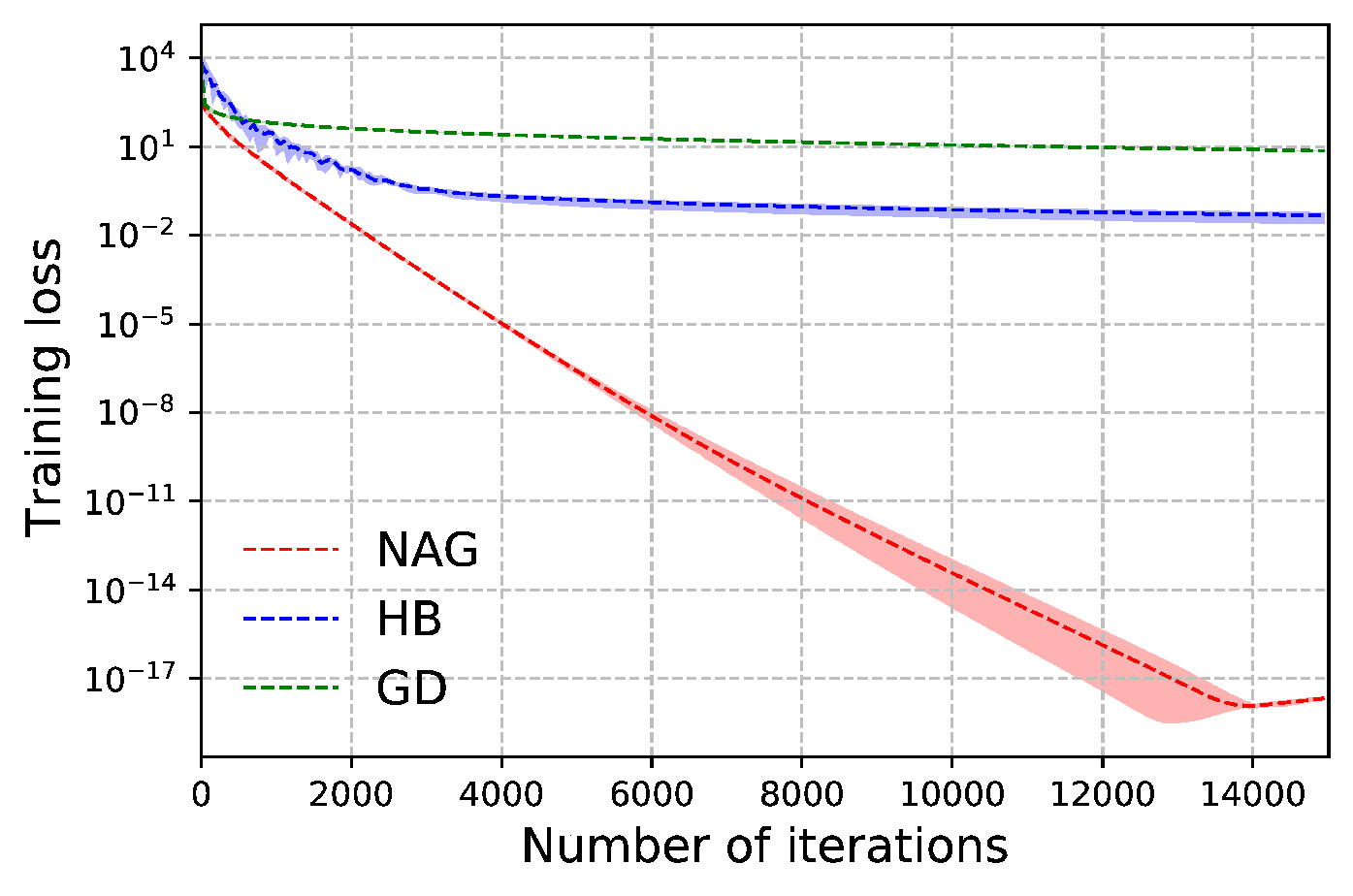}}\\
	\subfigure[CIFAR10]{\includegraphics[scale=0.47]{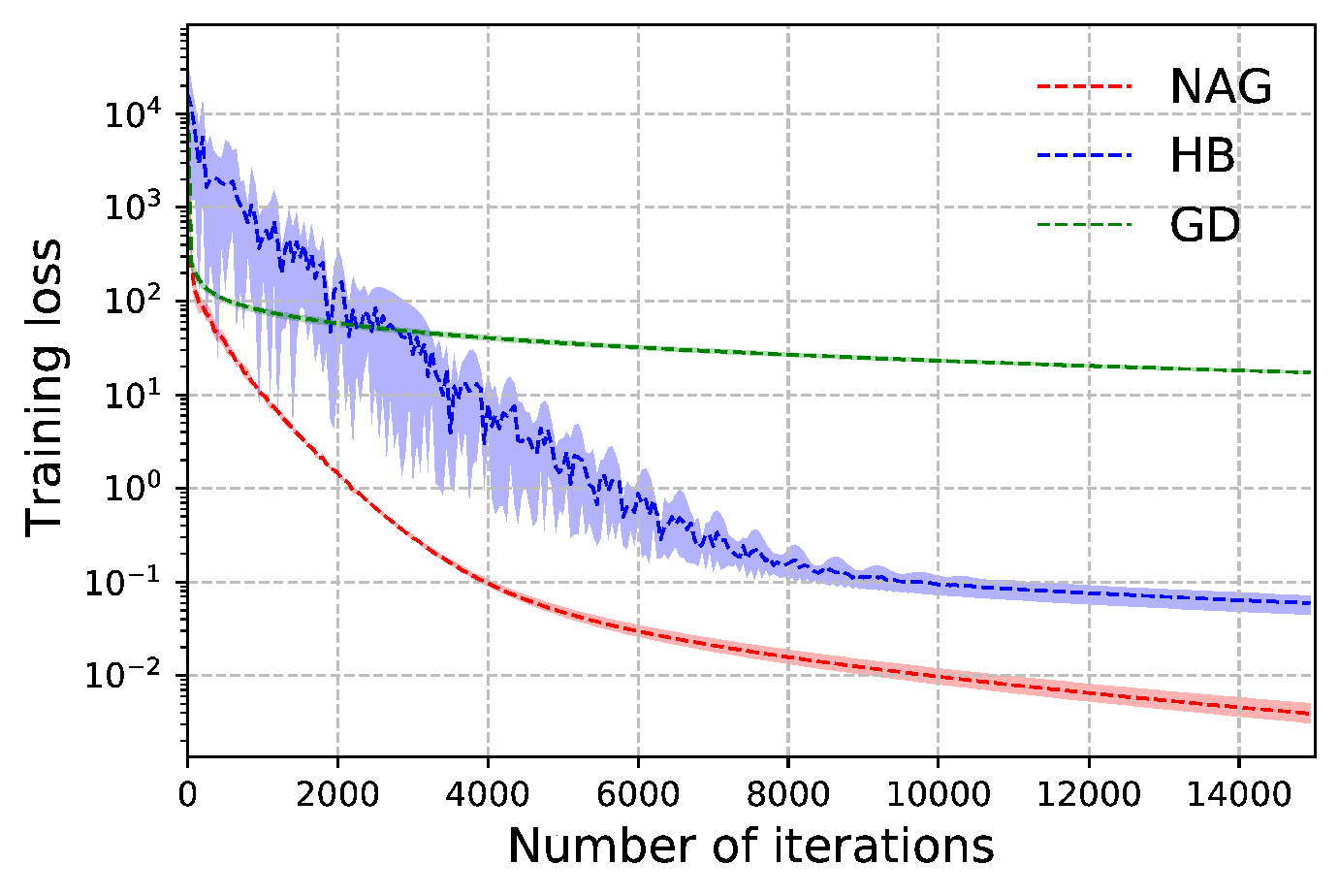}}
	\subfigure[ENERGY]{\includegraphics[scale=0.47]{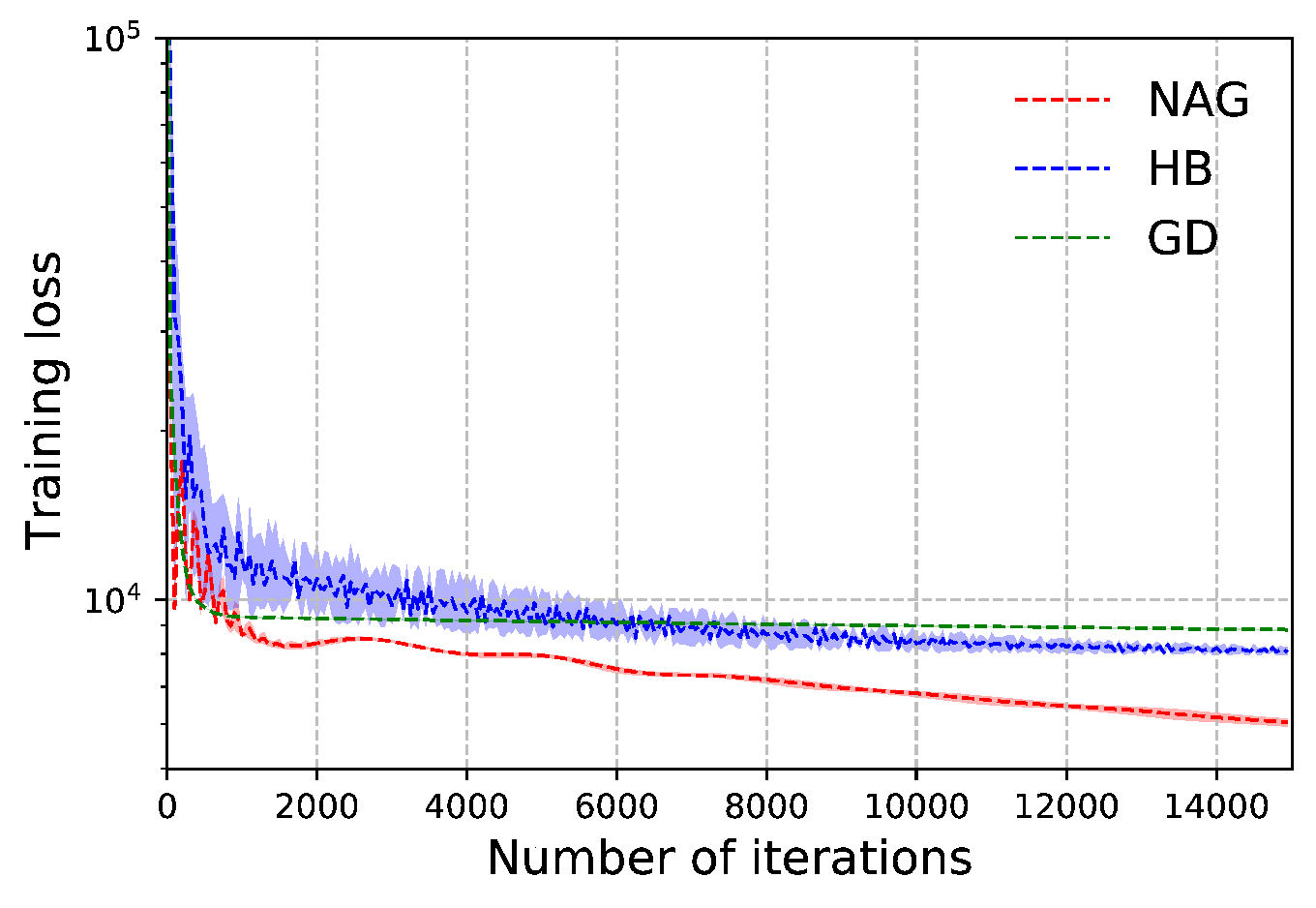}}\\
	\subfigure[HOUSING]{\includegraphics[scale=0.47]{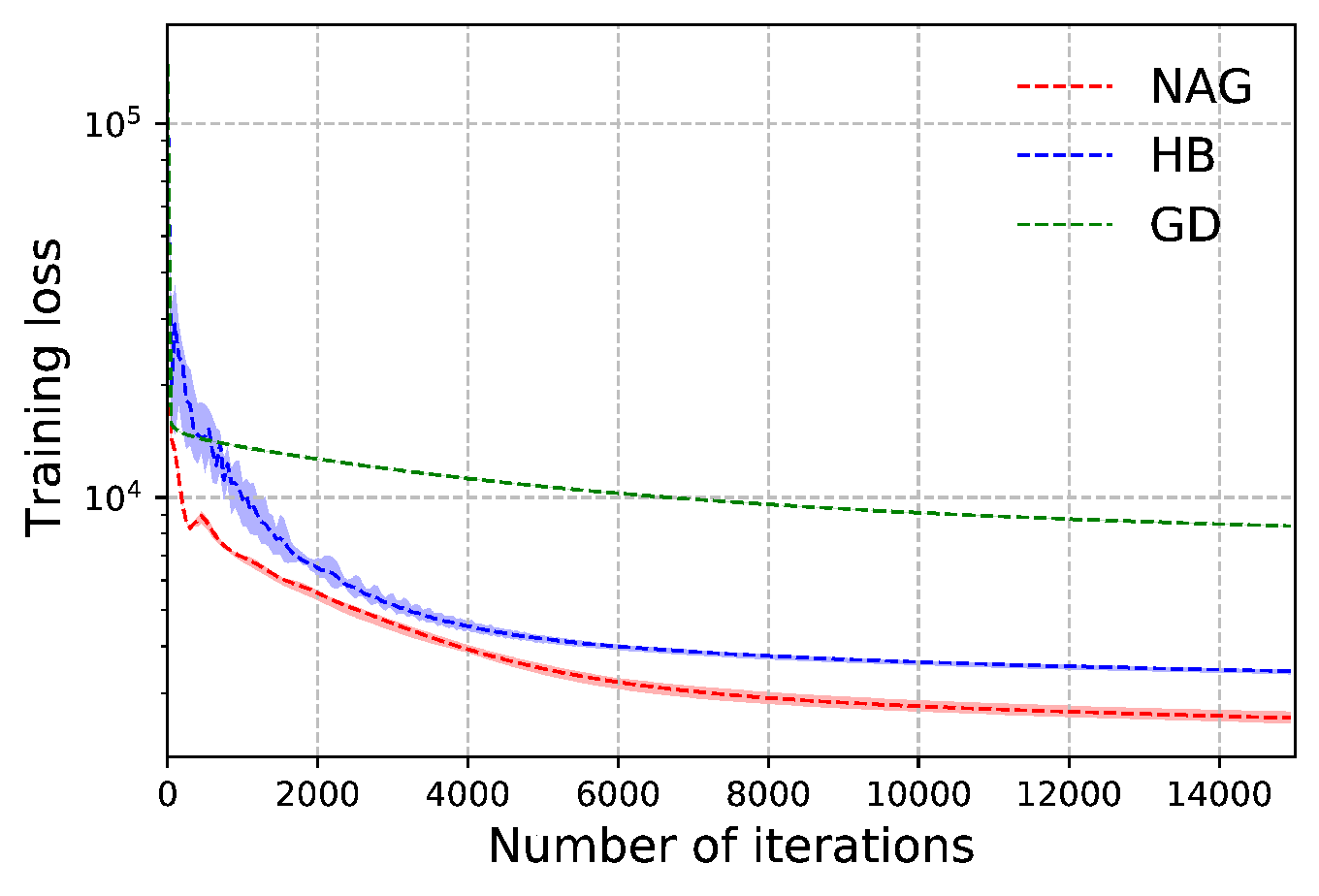}}
	\subfigure[YACHT]{\includegraphics[scale=0.47]{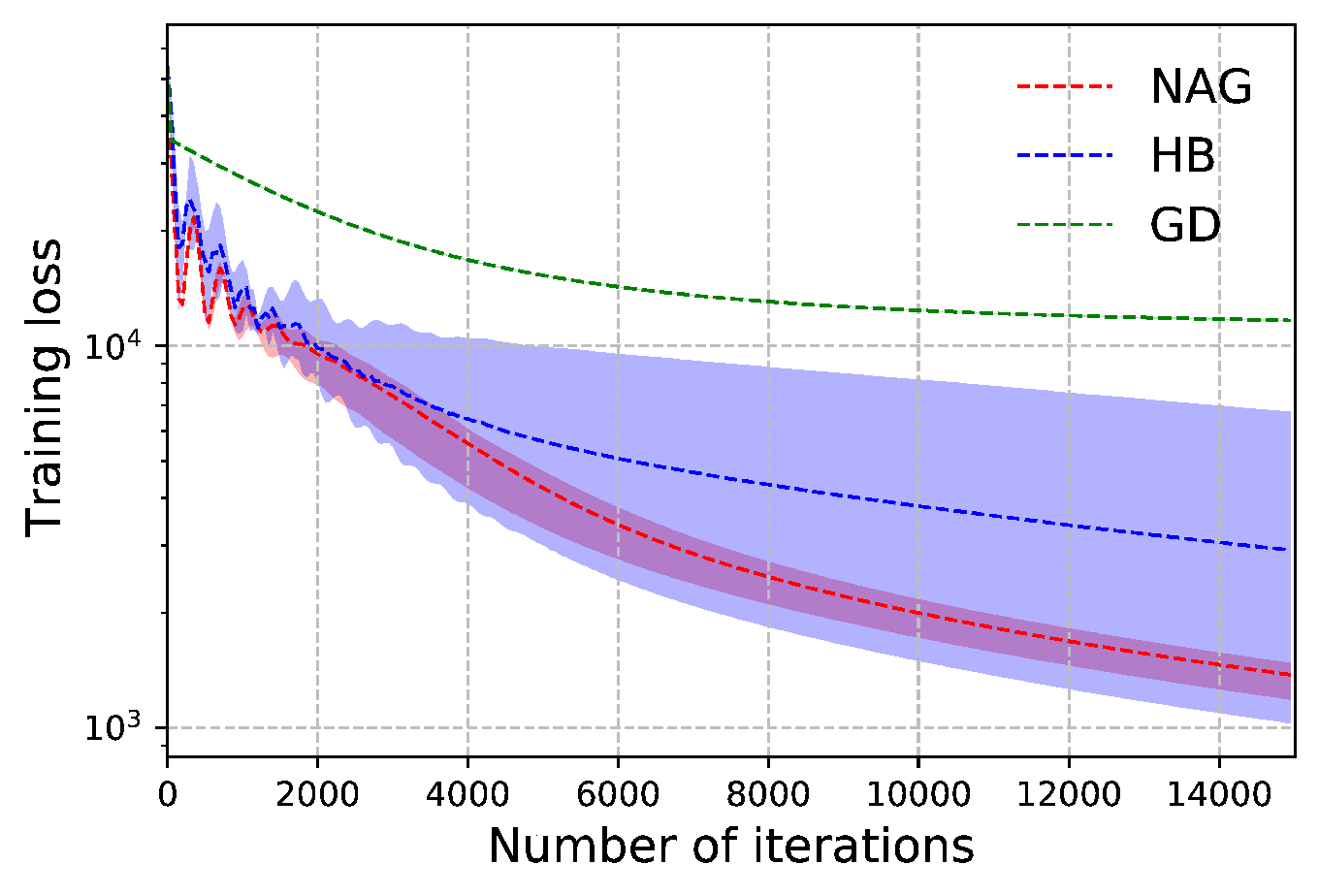}}
	\caption{Convergence comparison among GD, HB and NAG.
	}
	\label{Convergence}
\end{figure*}
\begin{figure*}[!t]
	\centering
	\subfigure[FMNIST]{\includegraphics[scale=0.47]{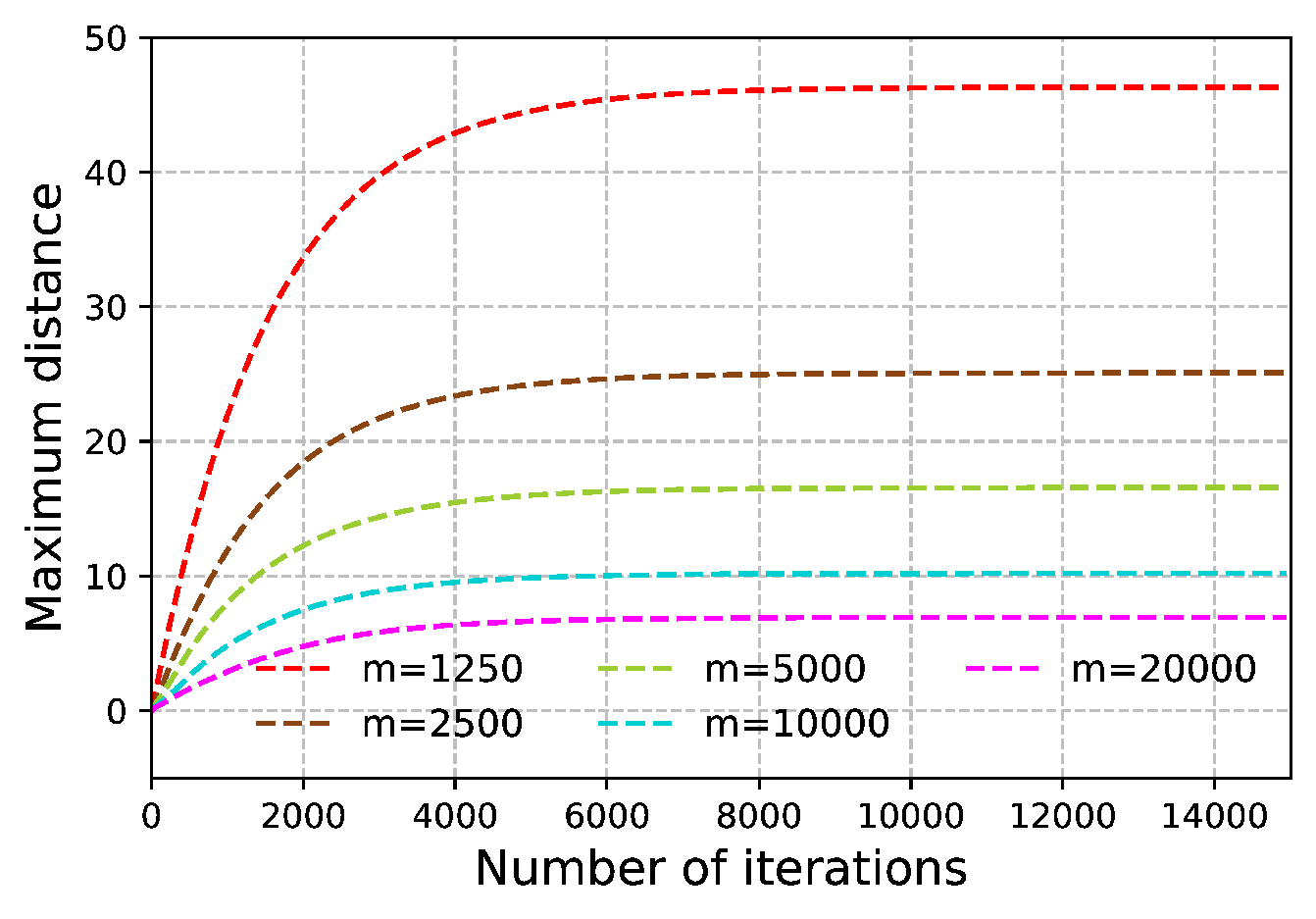}}
	\subfigure[MNIST]{\includegraphics[scale=0.47]{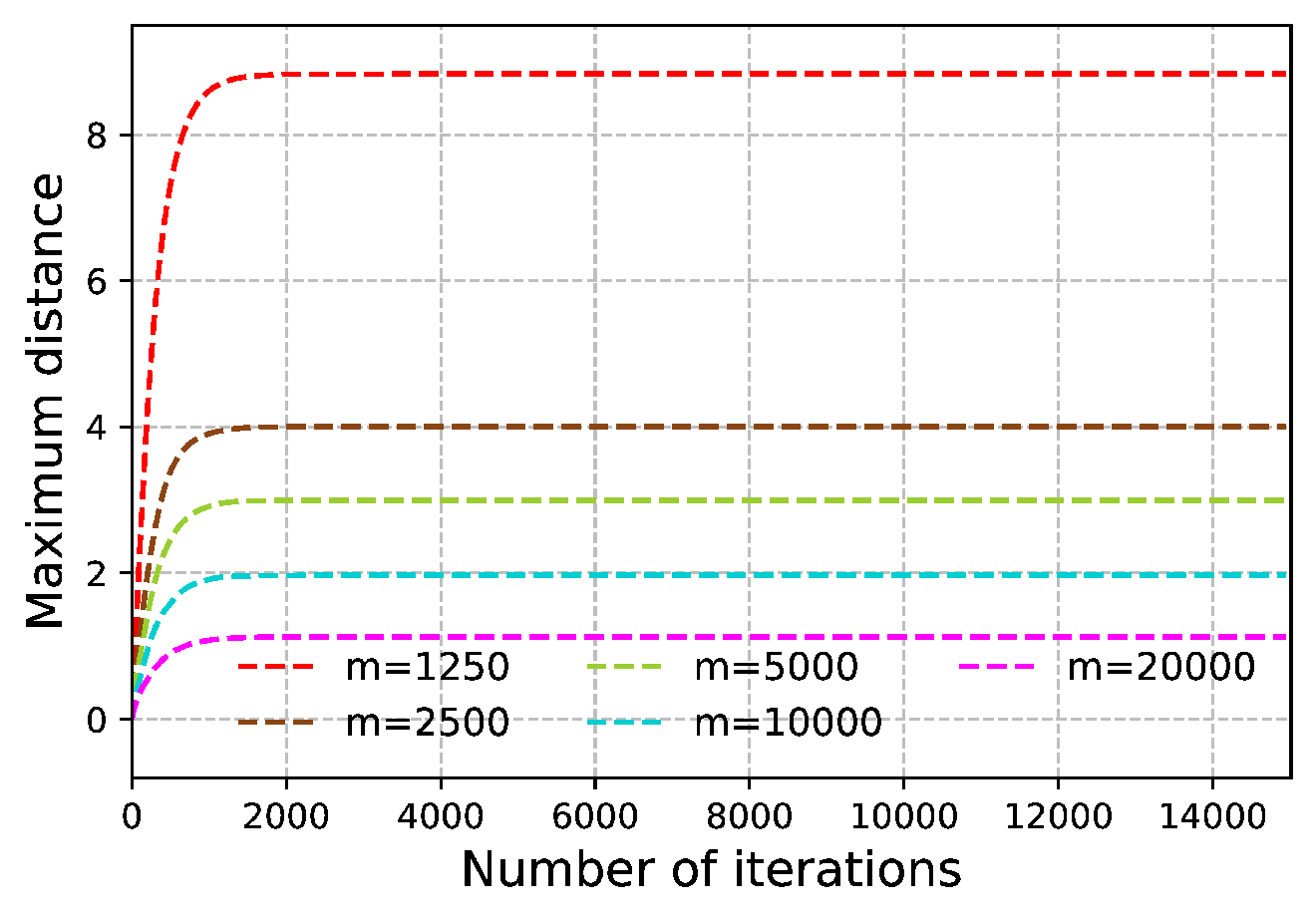}}\\
	\subfigure[CIFAR10]{\includegraphics[scale=0.47]{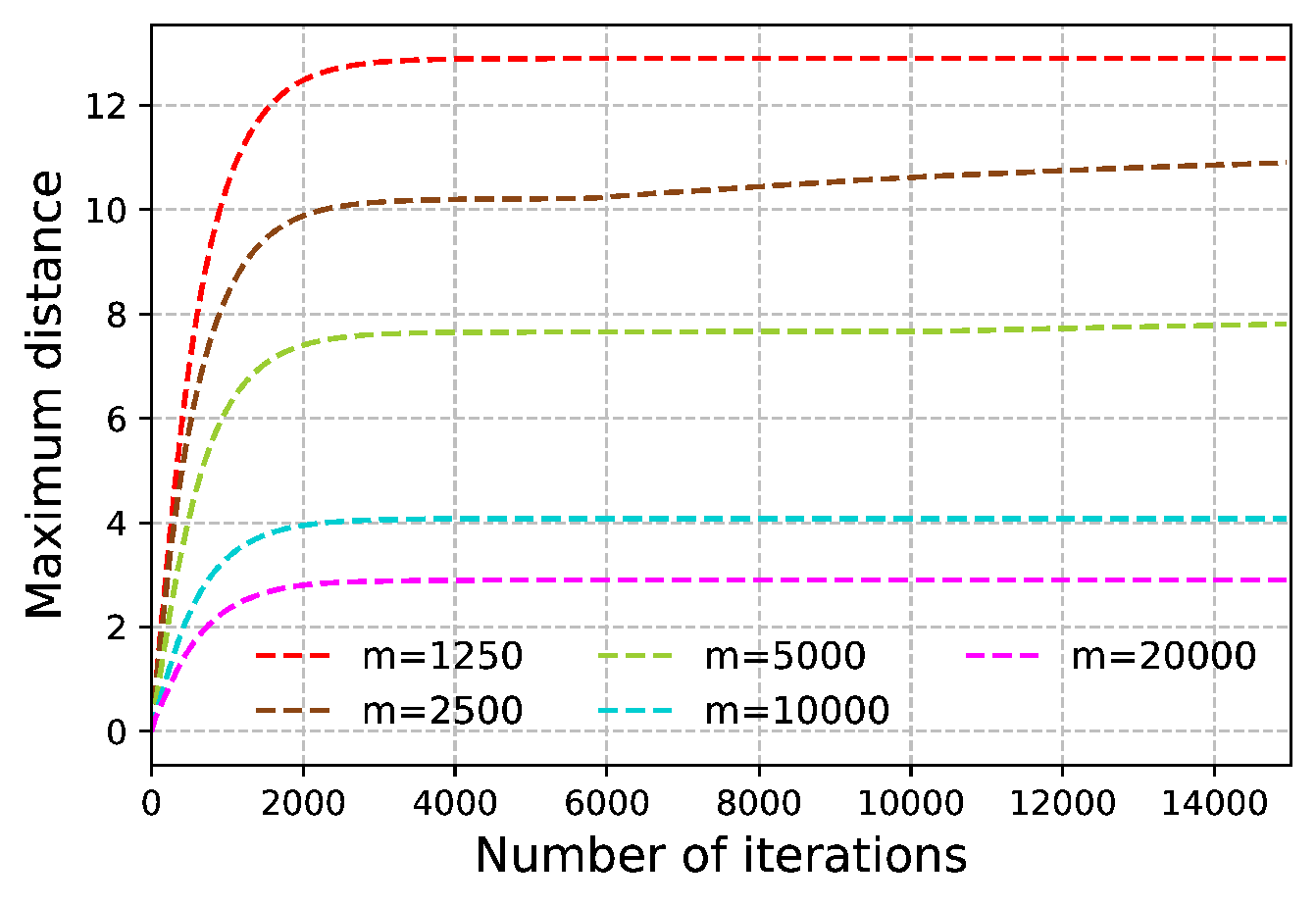}}
	\subfigure[ENERGY]{\includegraphics[scale=0.47]{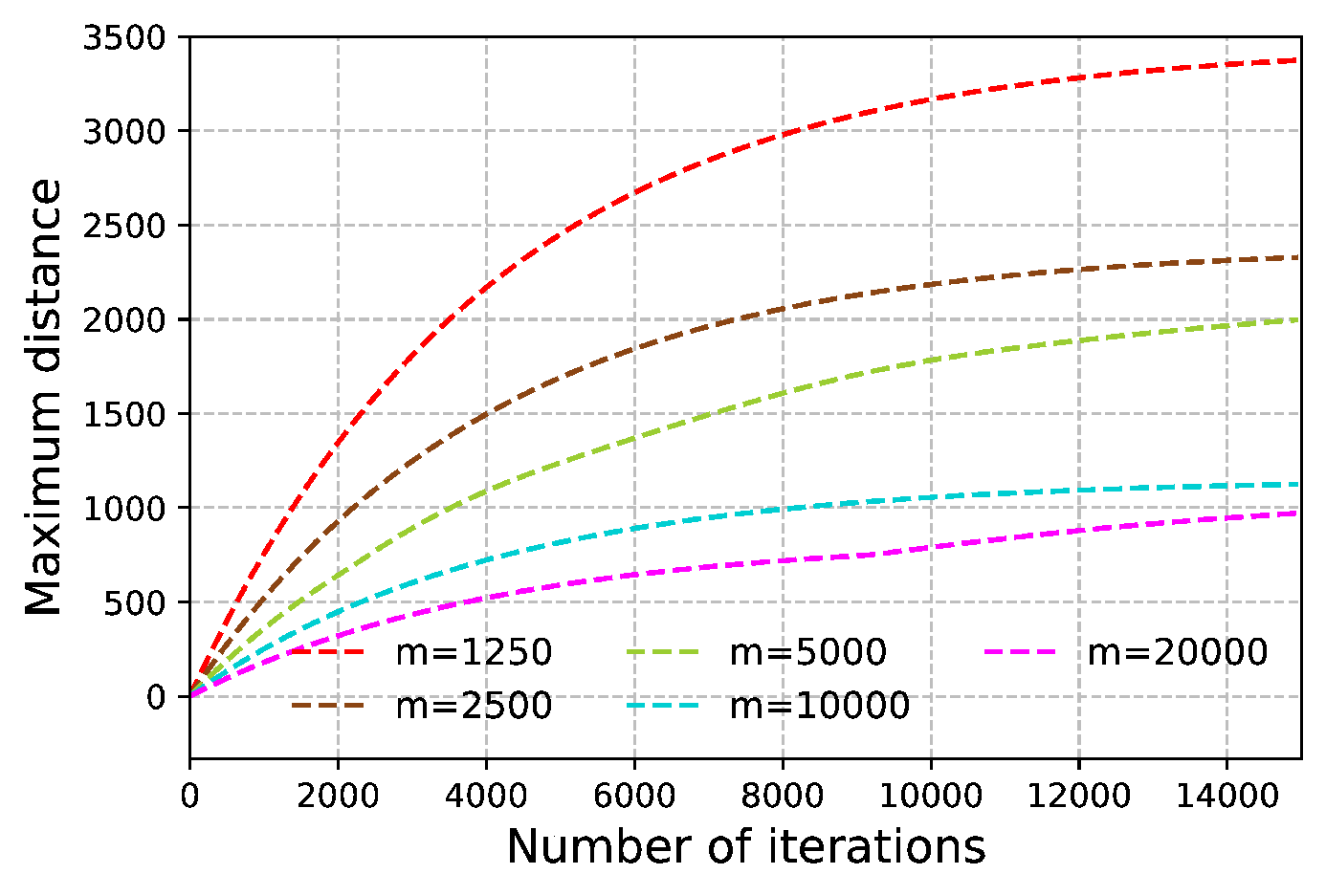}}\\
	\subfigure[HOUSING]{\includegraphics[scale=0.47]{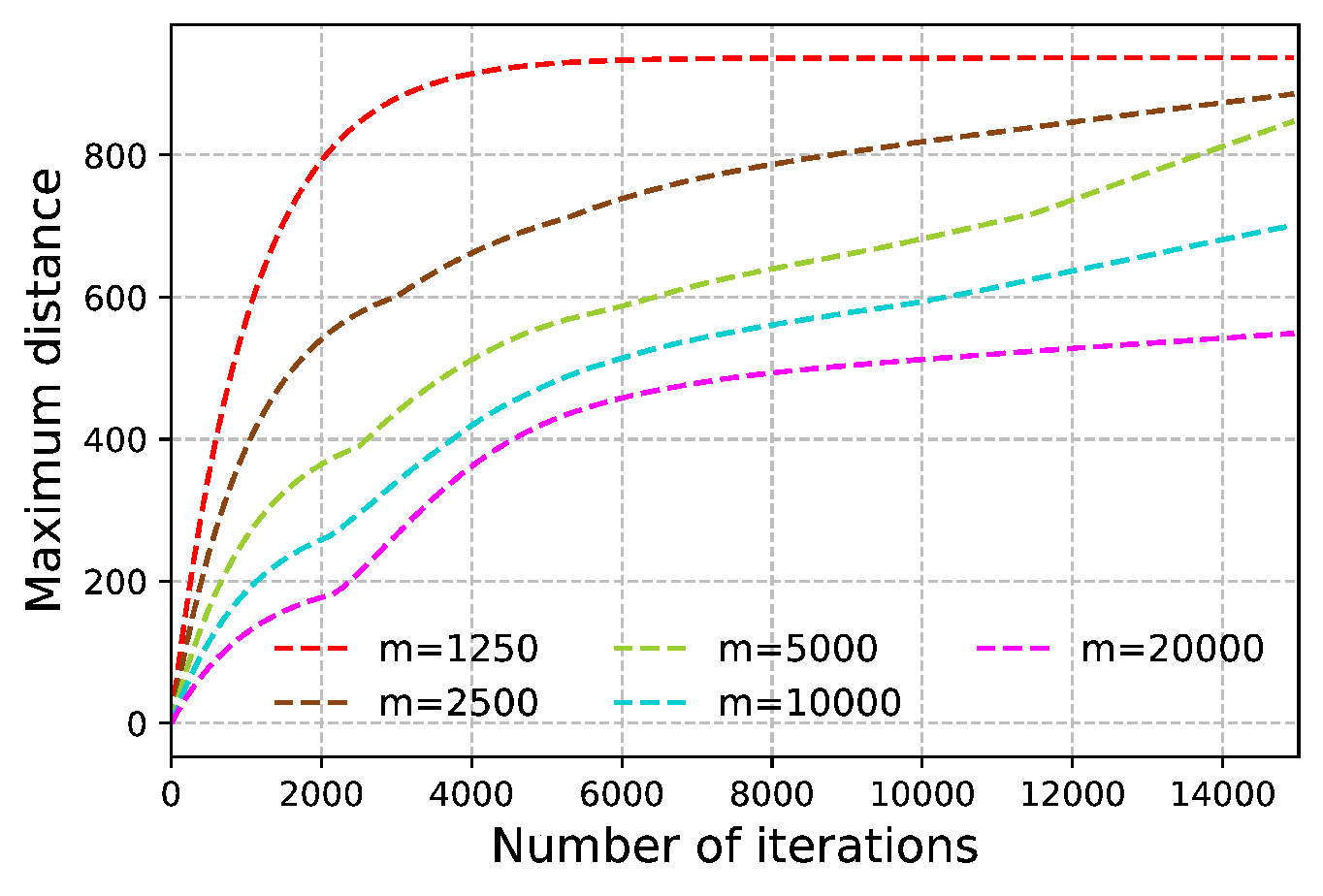}}
	\subfigure[YACHT]{\includegraphics[scale=0.47]{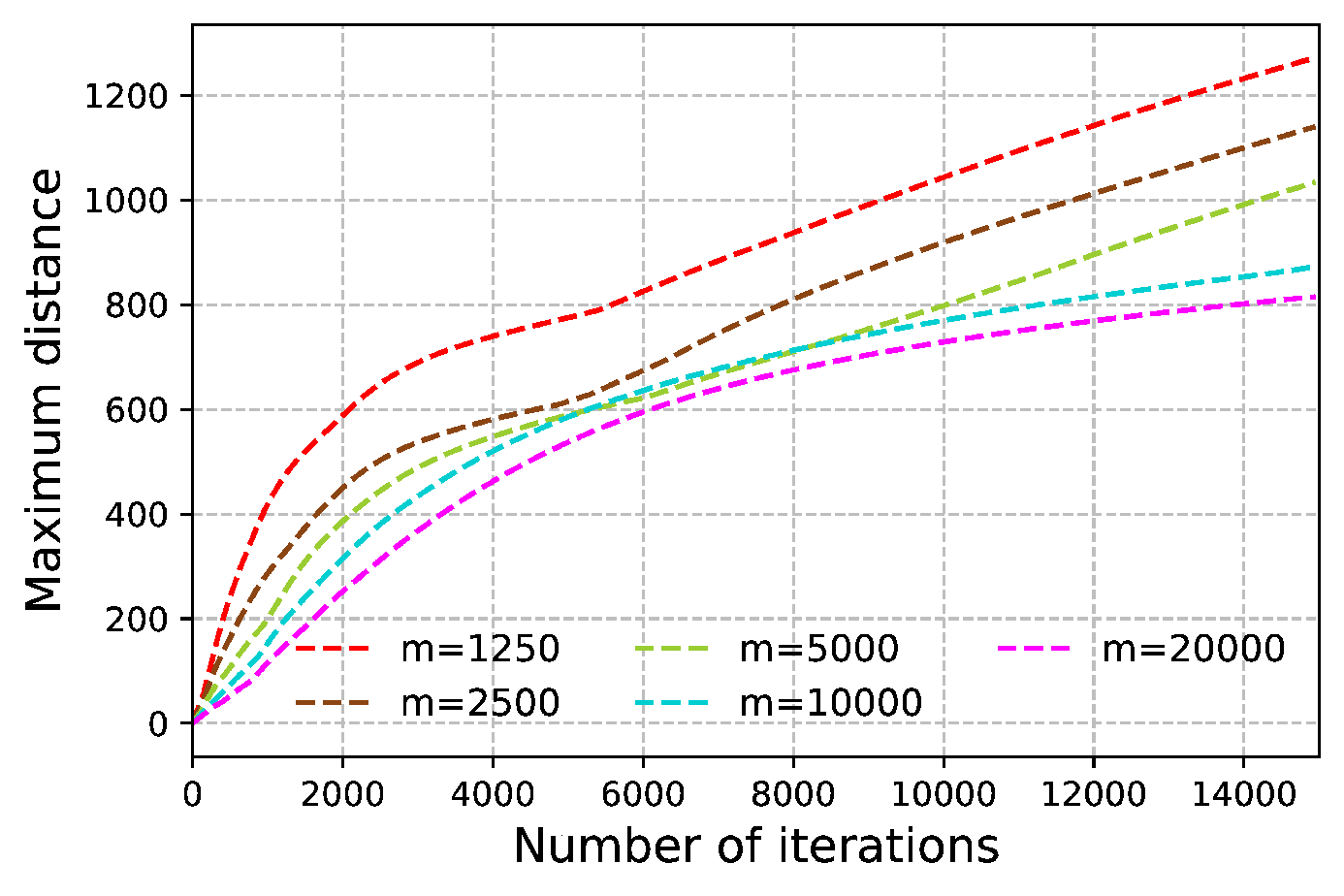}}
	\caption{Maximum distance from initialization comparison for NAG with different width $m$.
	}
	\label{relative_distance}
\end{figure*}

\begin{figure*}[!t]
	\centering
	\subfigure[FMNIST]{\includegraphics[scale=0.47]{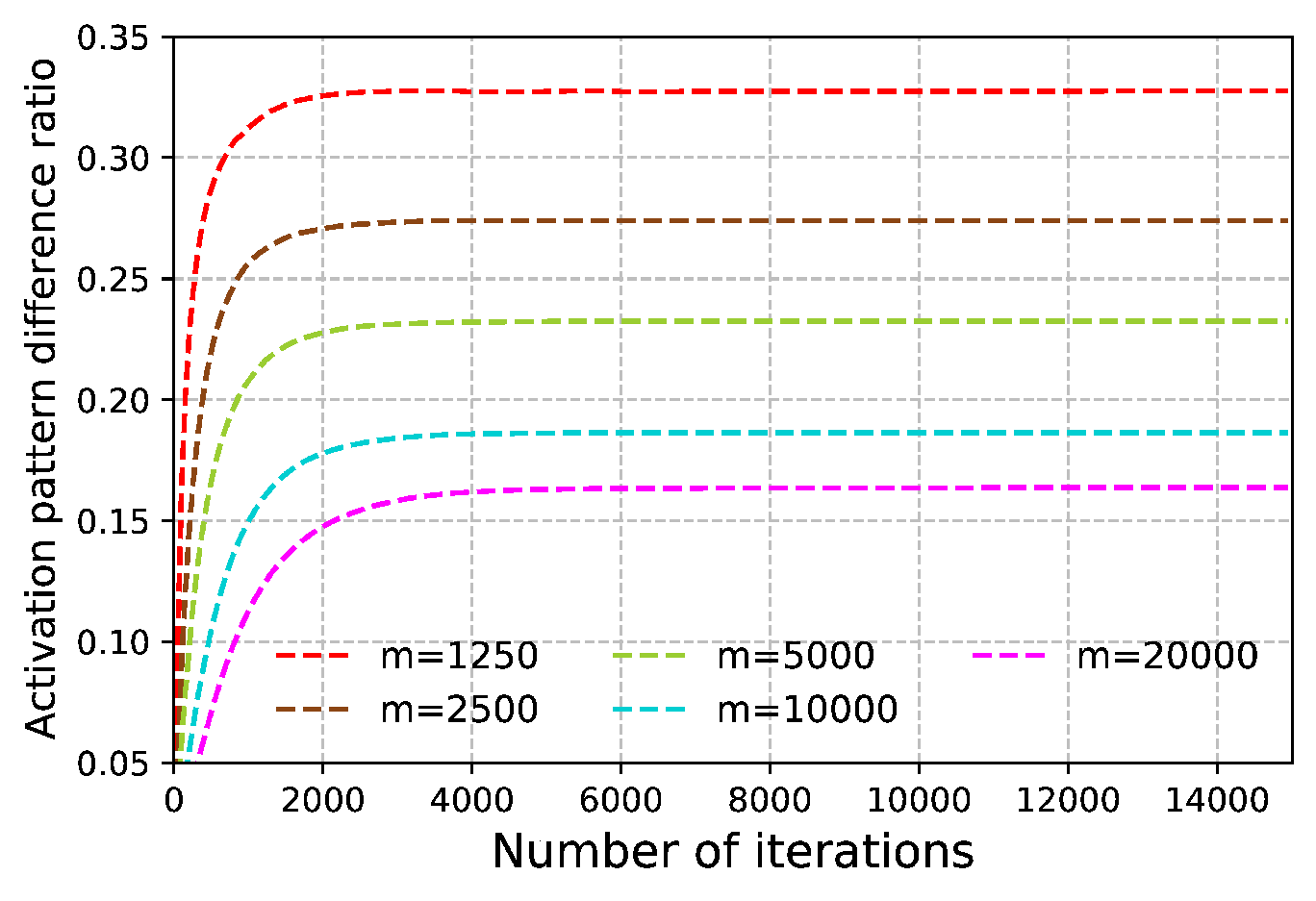}}
	\subfigure[MNIST]{\includegraphics[scale=0.47]{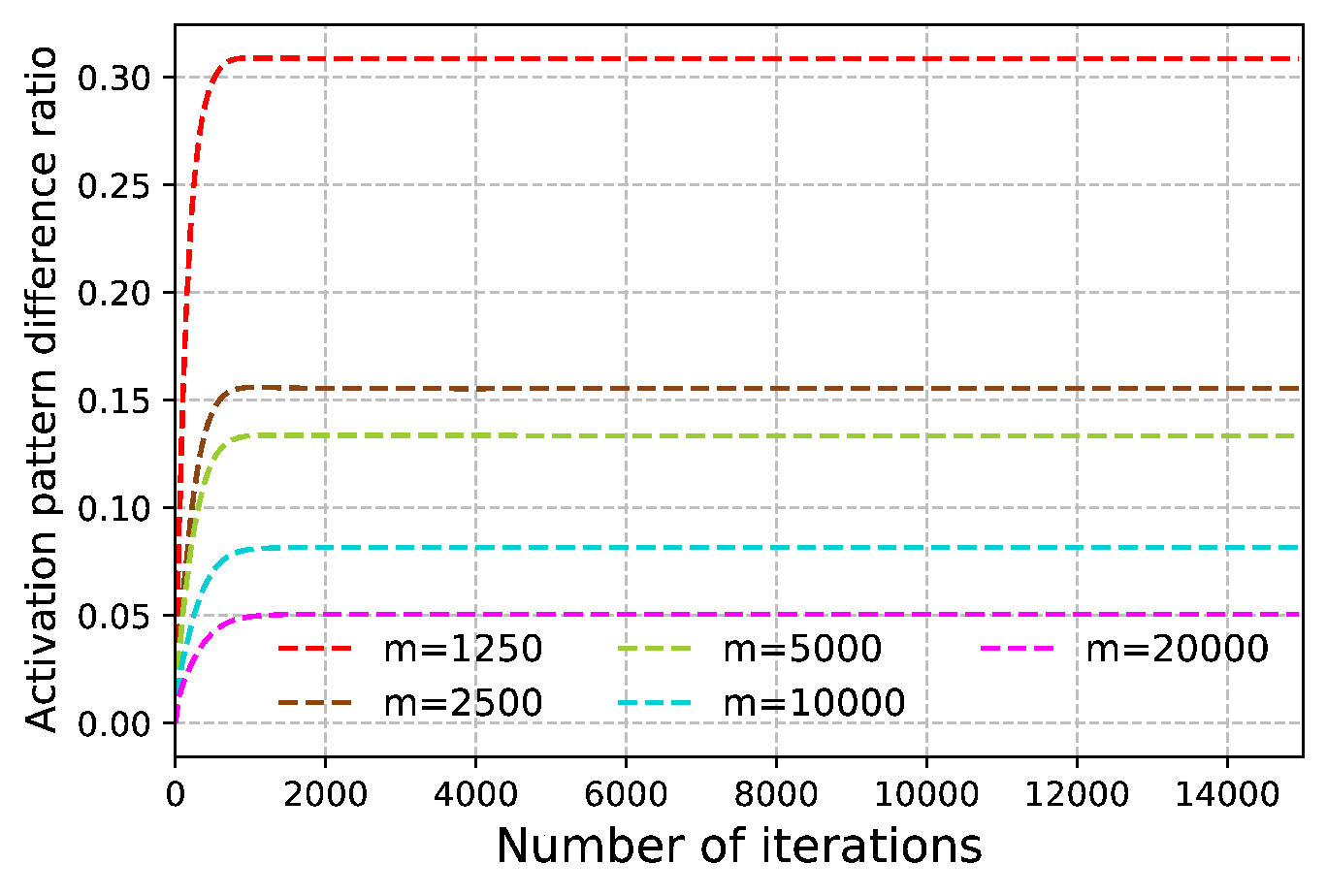}}\\
	\subfigure[CIFAR10]{\includegraphics[scale=0.47]{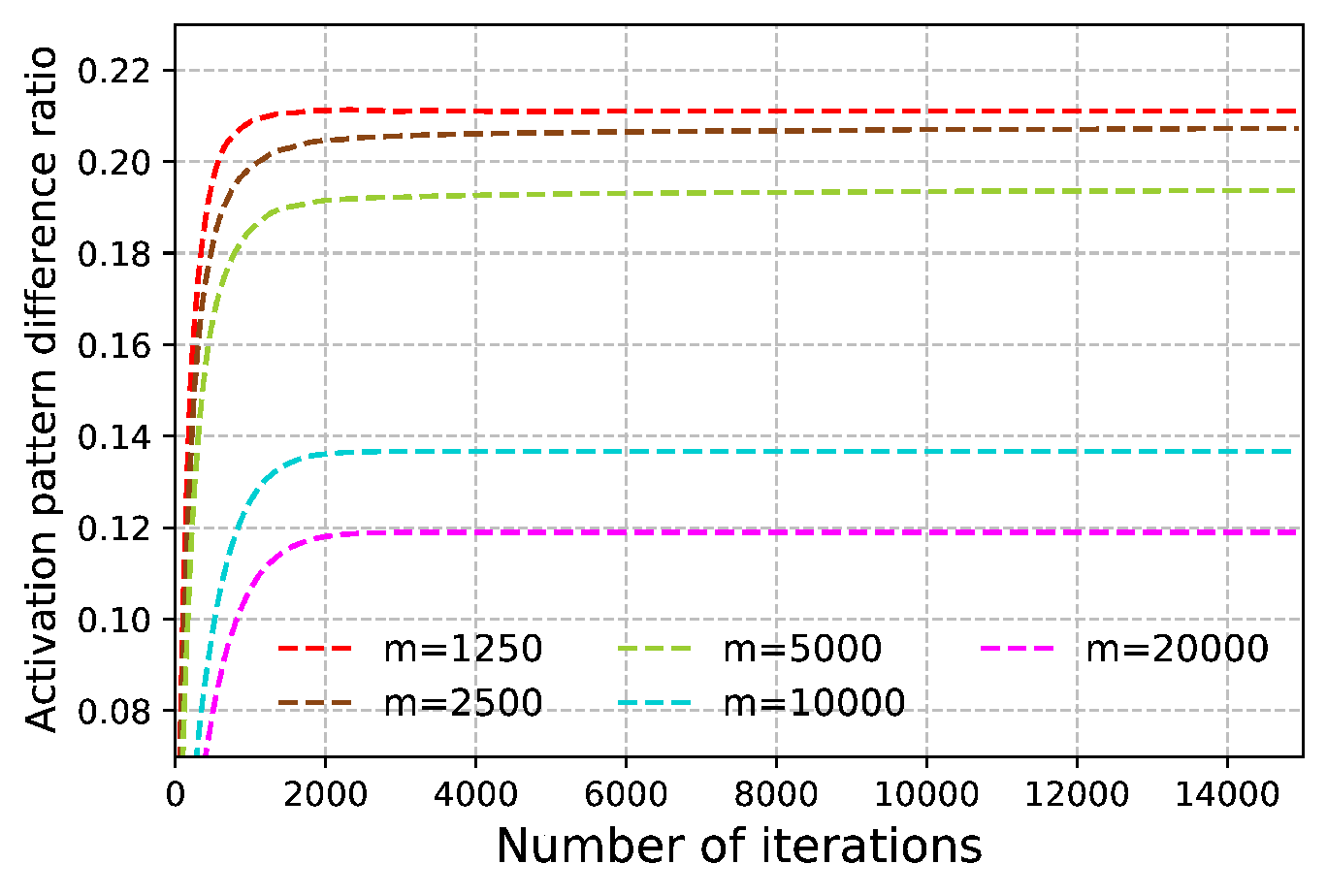}}
	\subfigure[ENERGY]{\includegraphics[scale=0.47]{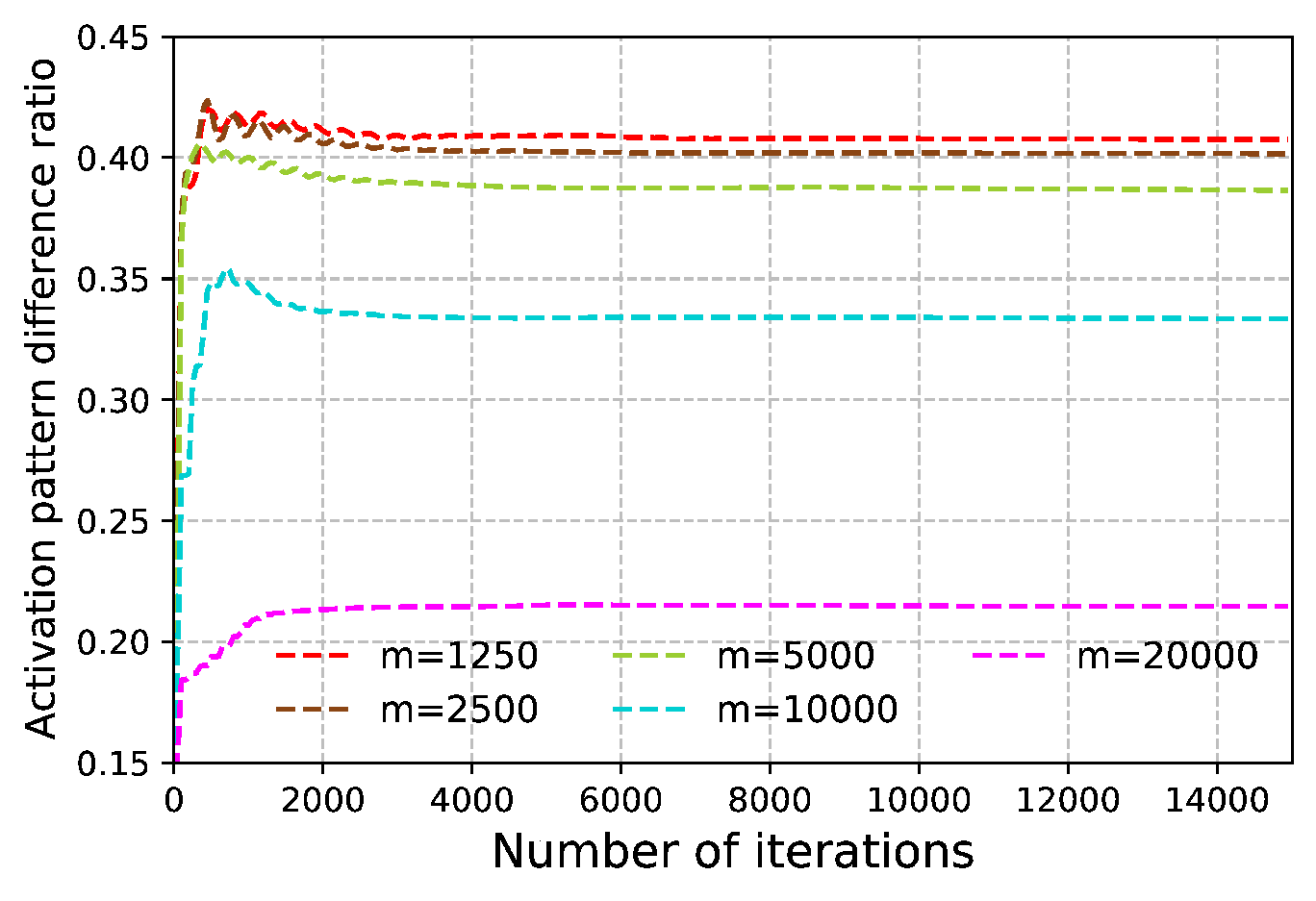}}\\
	\subfigure[HOUSING]{\includegraphics[scale=0.47]{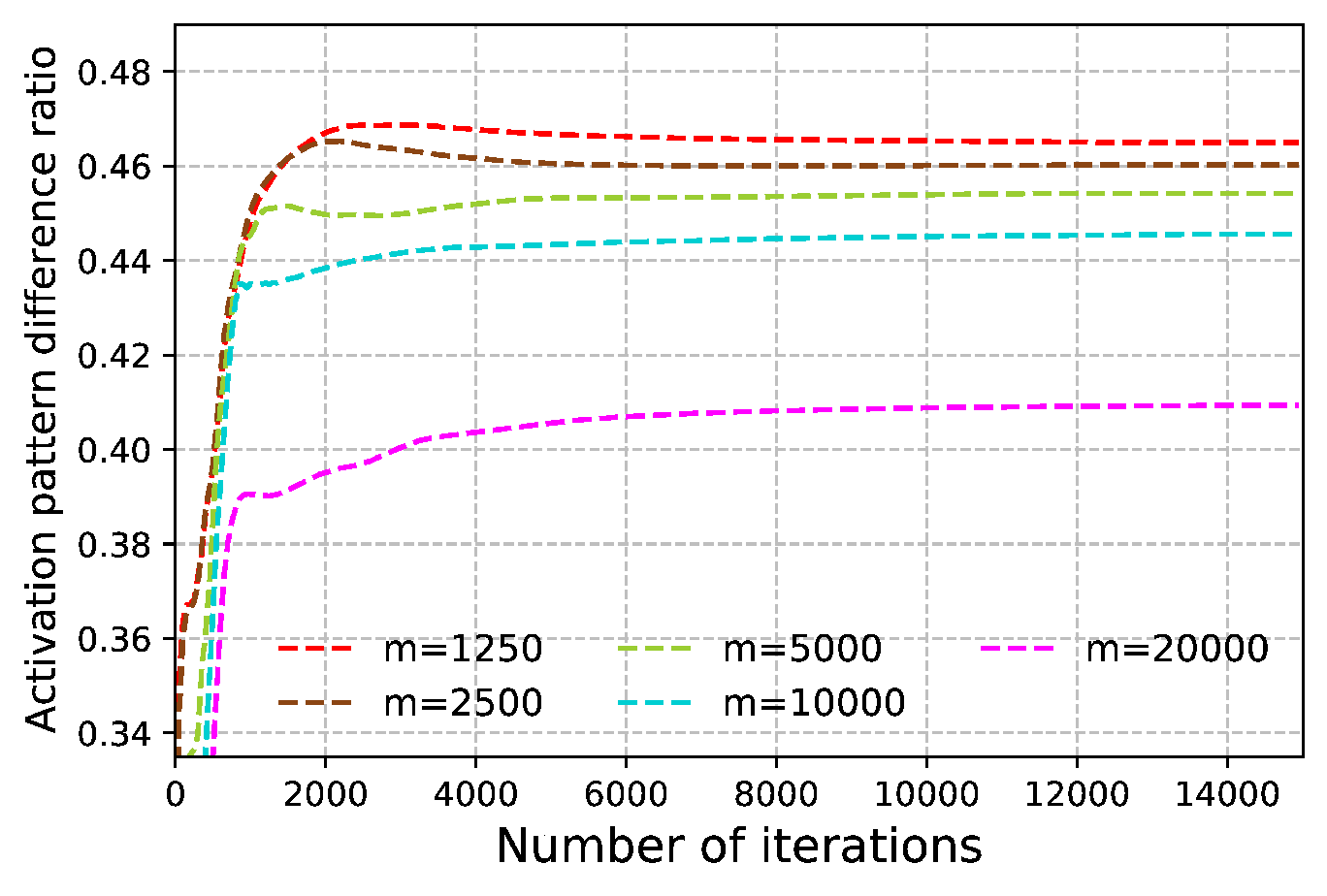}}
	\subfigure[YACHT]{\includegraphics[scale=0.47]{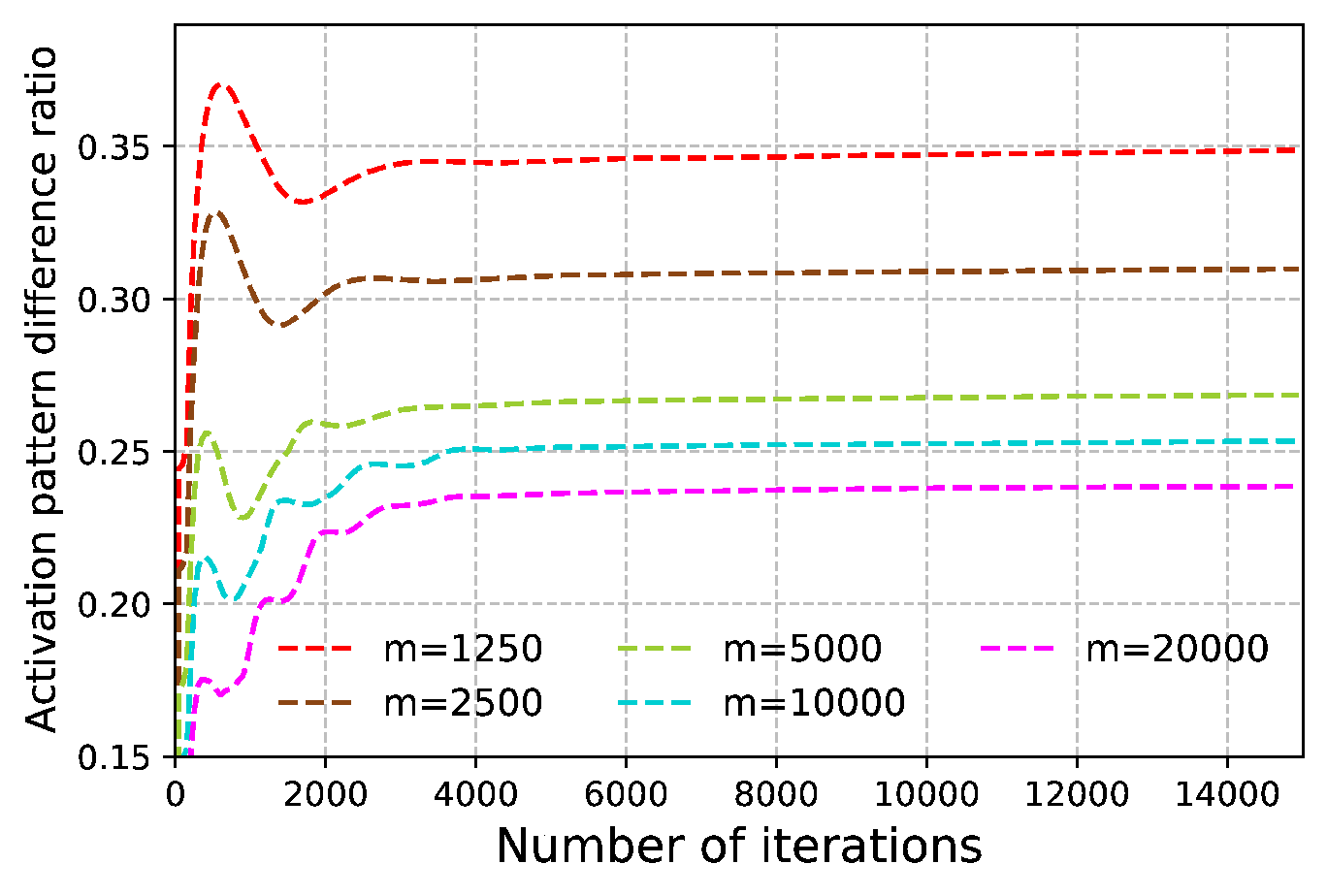}}
	\caption{Activation pattern difference ratio comparison for NAG with different width $m$. 
	}
	\label{activation_pattern}
\end{figure*}

\begin{figure*}[!t]
	\centering
	\subfigure[FMNIST]{\includegraphics[scale=0.47]{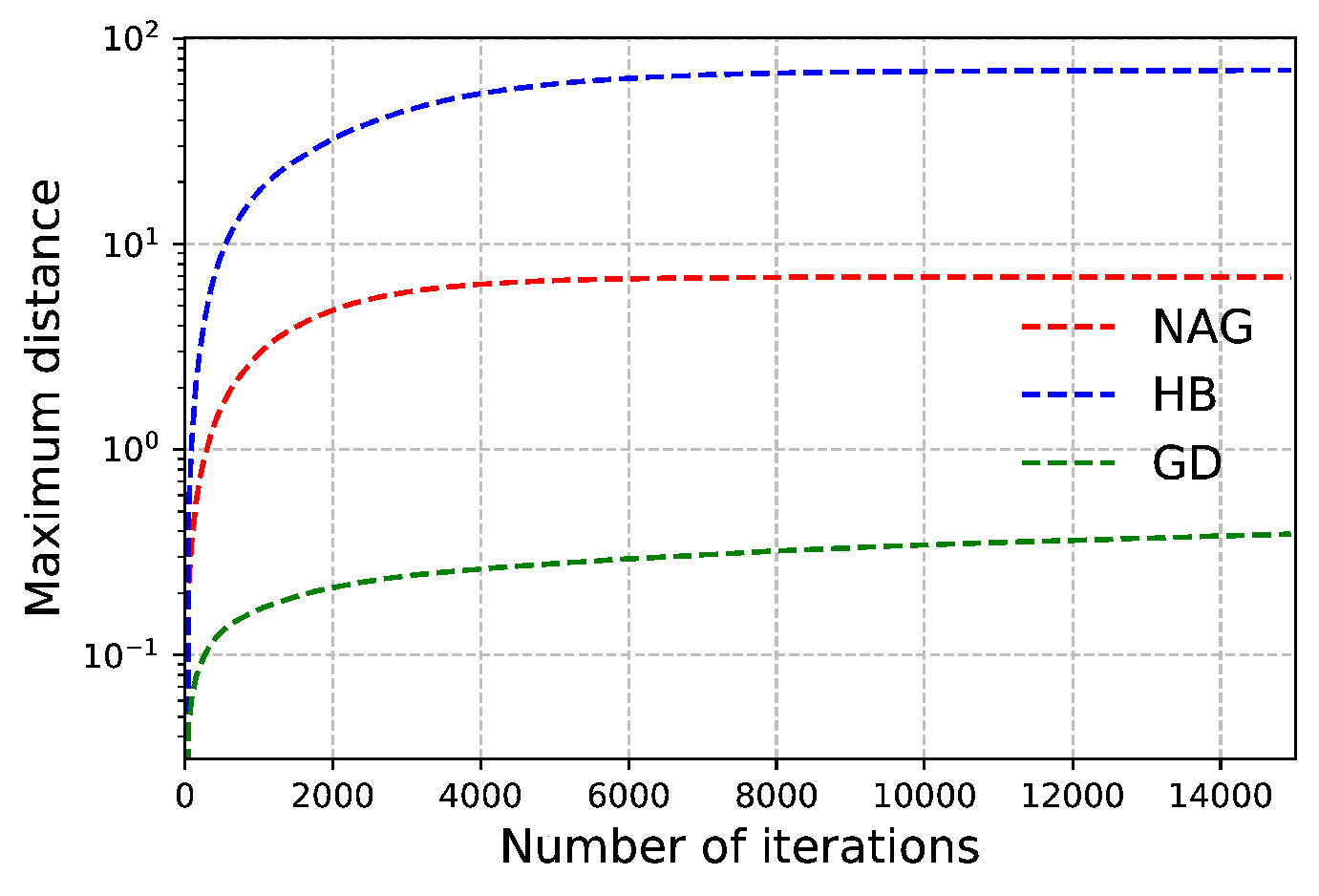}}
	\subfigure[MNIST]{\includegraphics[scale=0.47]{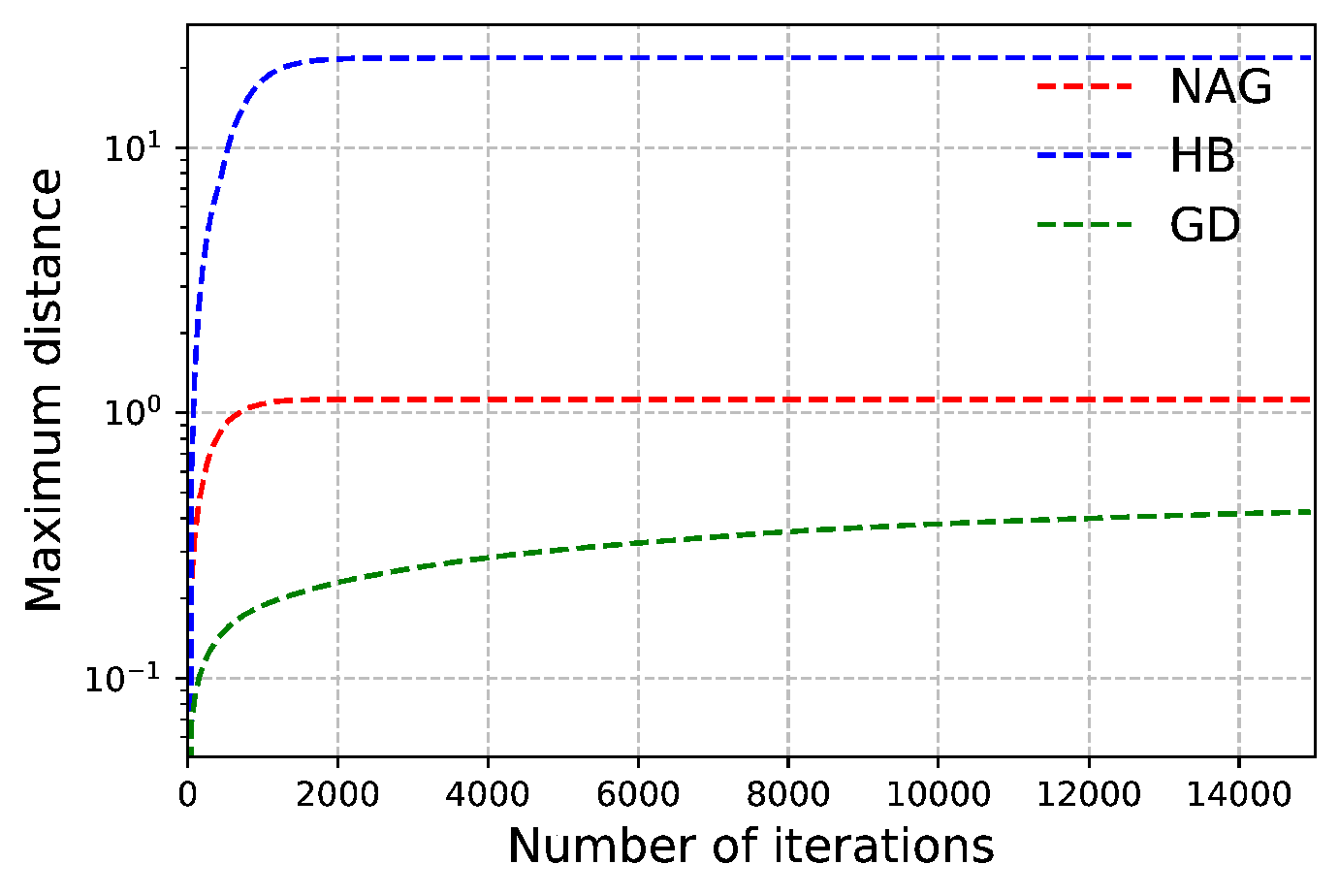}}\\
	\subfigure[CIFAR10]{\includegraphics[scale=0.47]{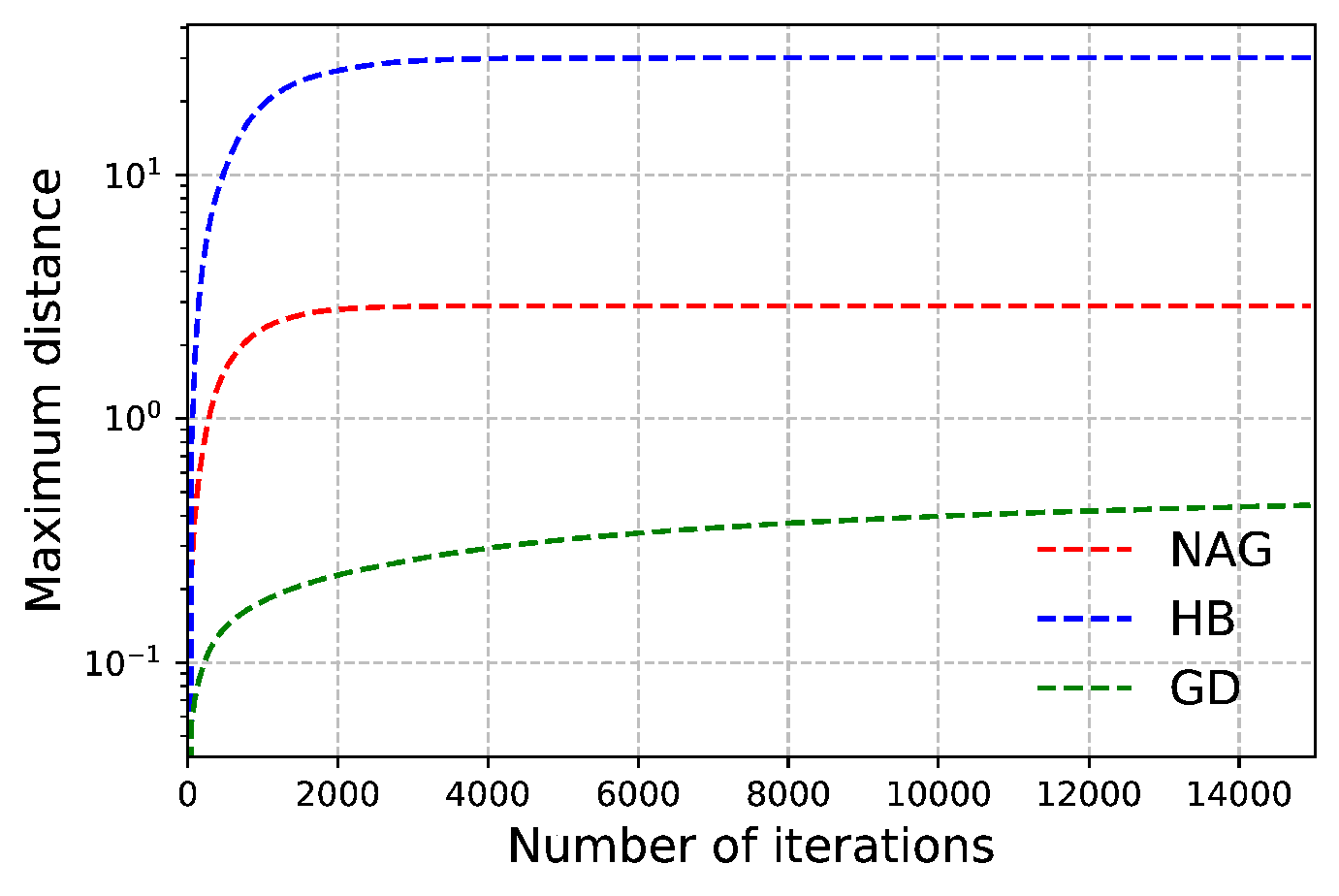}}
	\subfigure[ENERGY]{\includegraphics[scale=0.47]{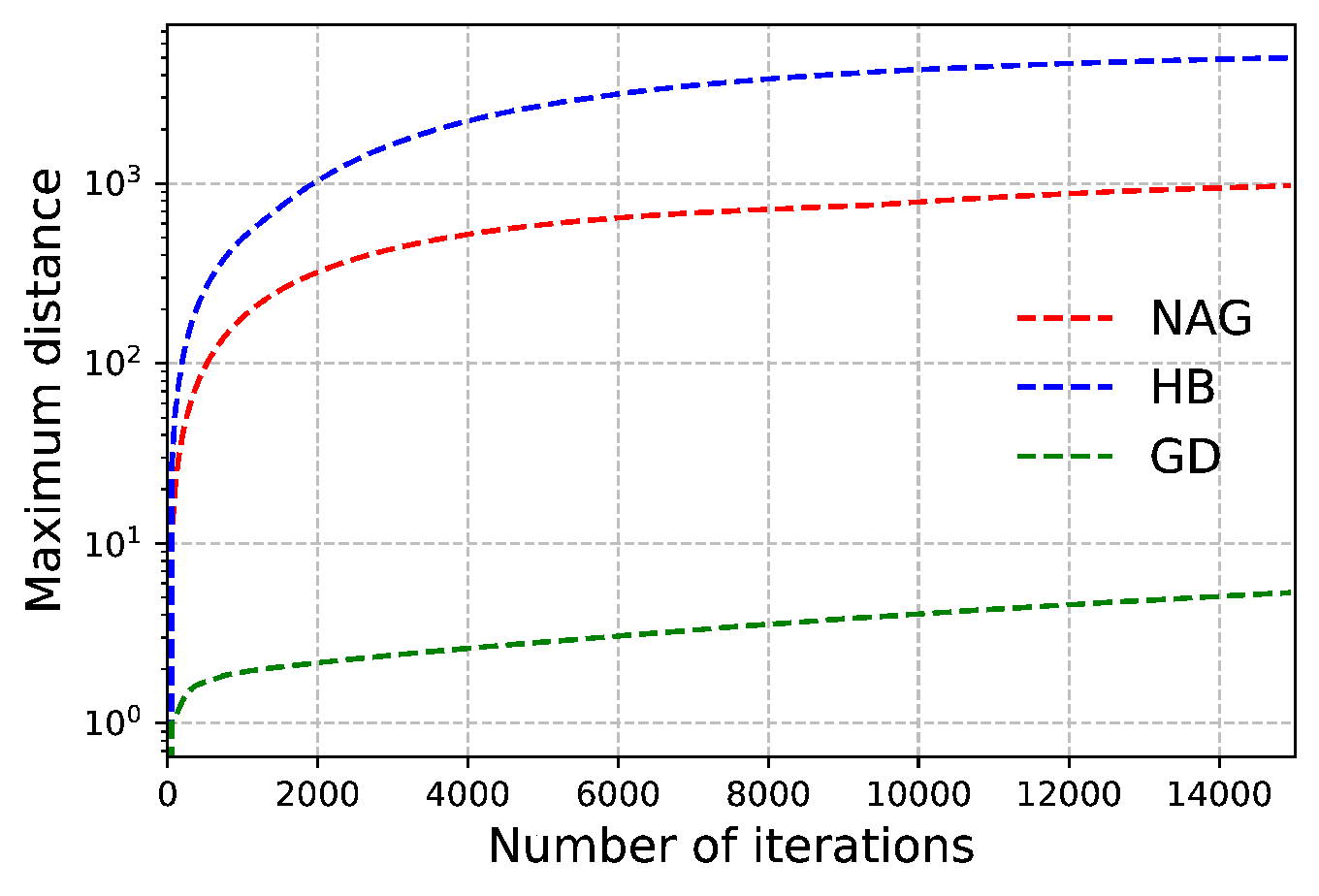}}\\
	\subfigure[HOUSING]{\includegraphics[scale=0.47]{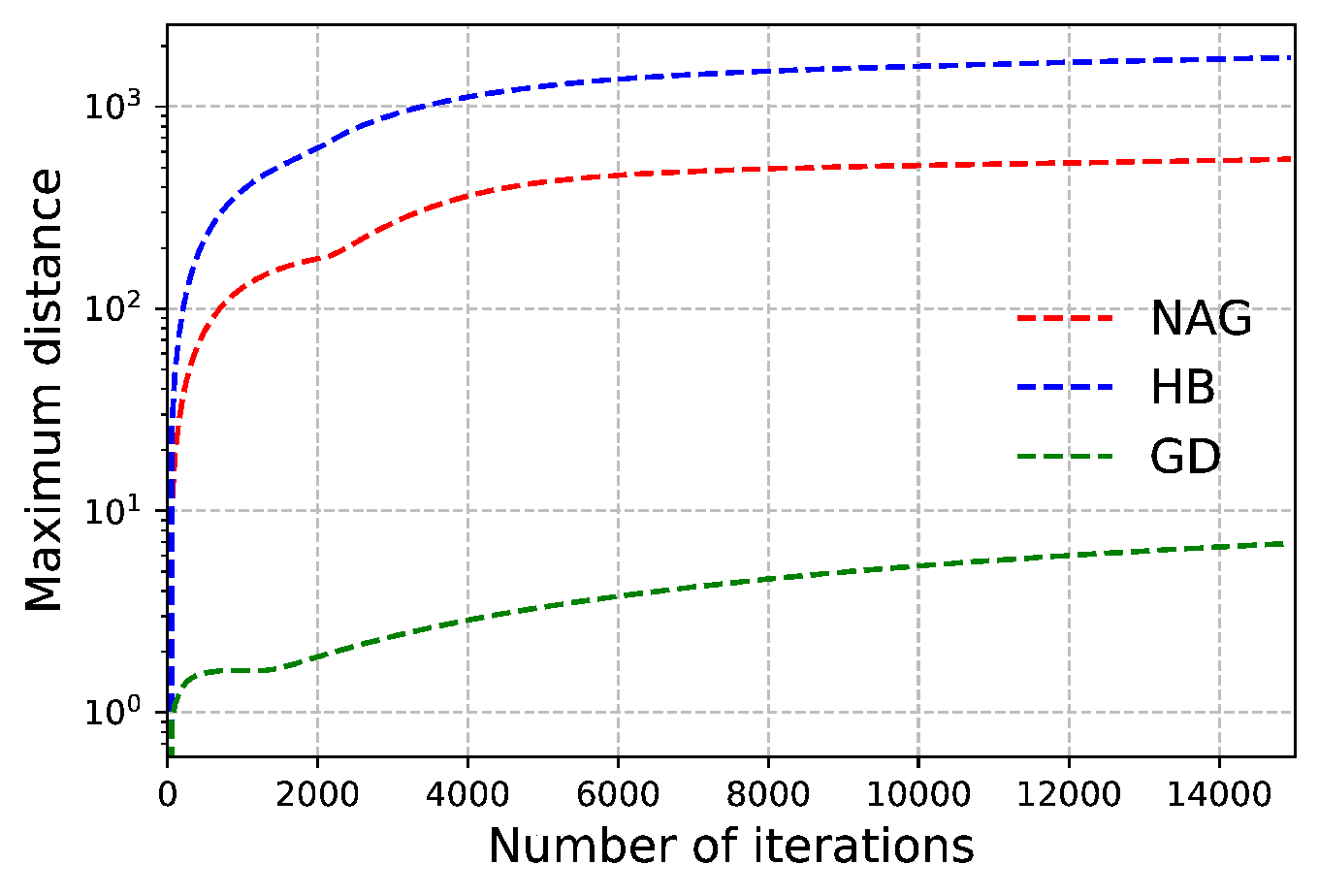}}
	\subfigure[YACHT]{\includegraphics[scale=0.47]{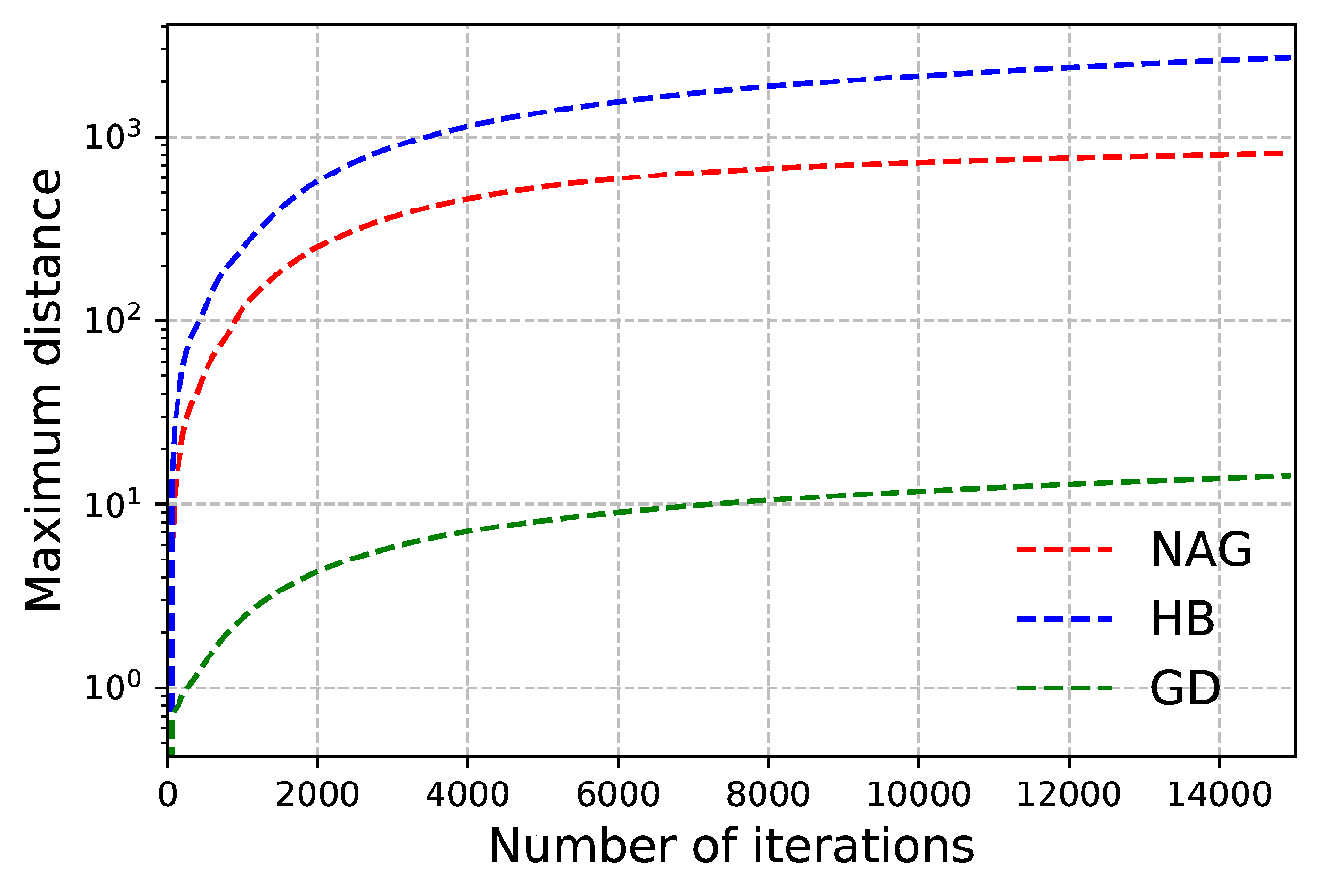}}
	\caption{Maximum distance from initialization comparison among GD, HB and NAG with width $m=20000$.
	}
	\label{relative_distance_d}
\end{figure*}

\begin{figure*}[!t]
	\centering
	\subfigure[FMNIST]{\includegraphics[scale=0.47]{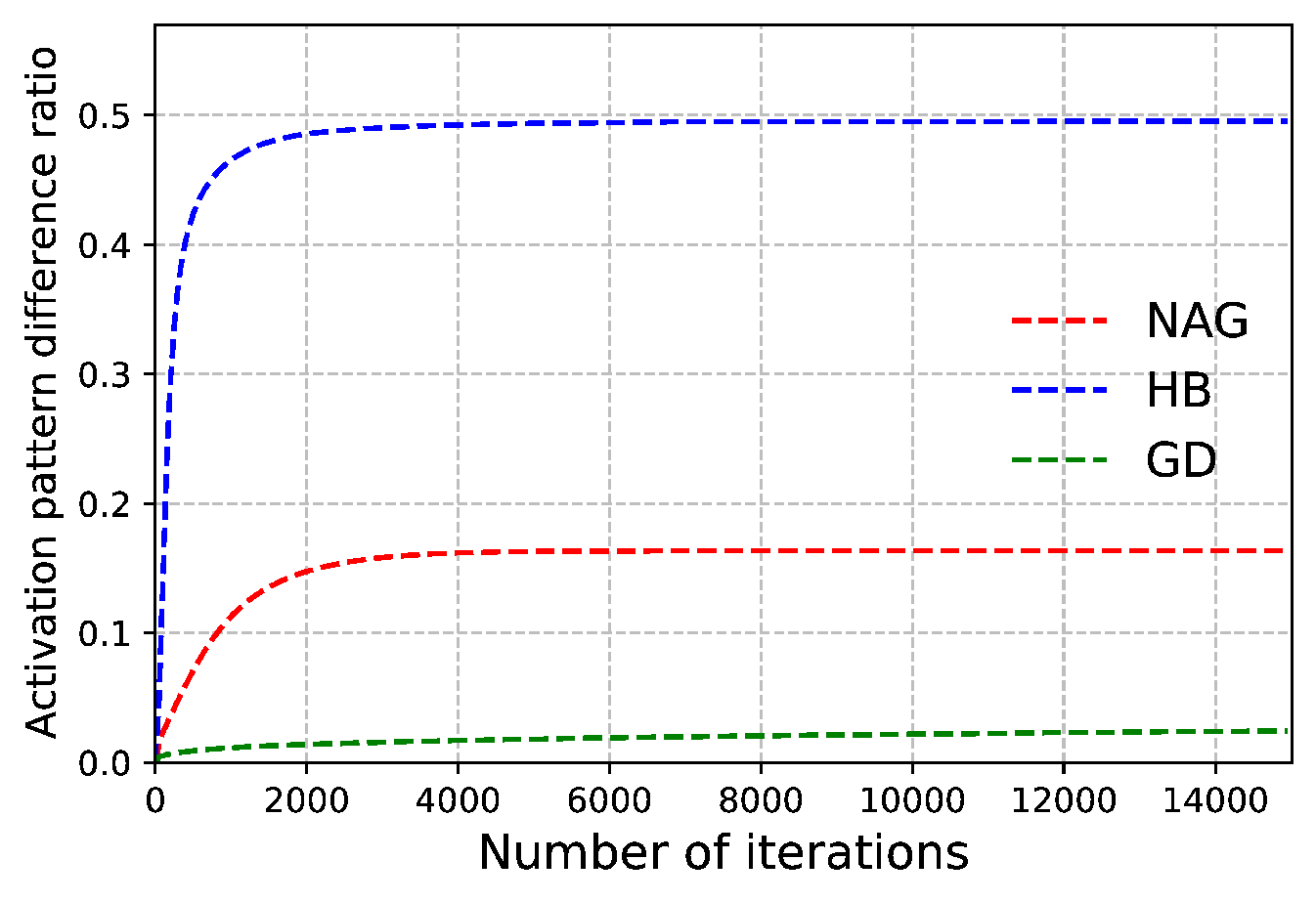}}
	\subfigure[MNIST]{\includegraphics[scale=0.47]{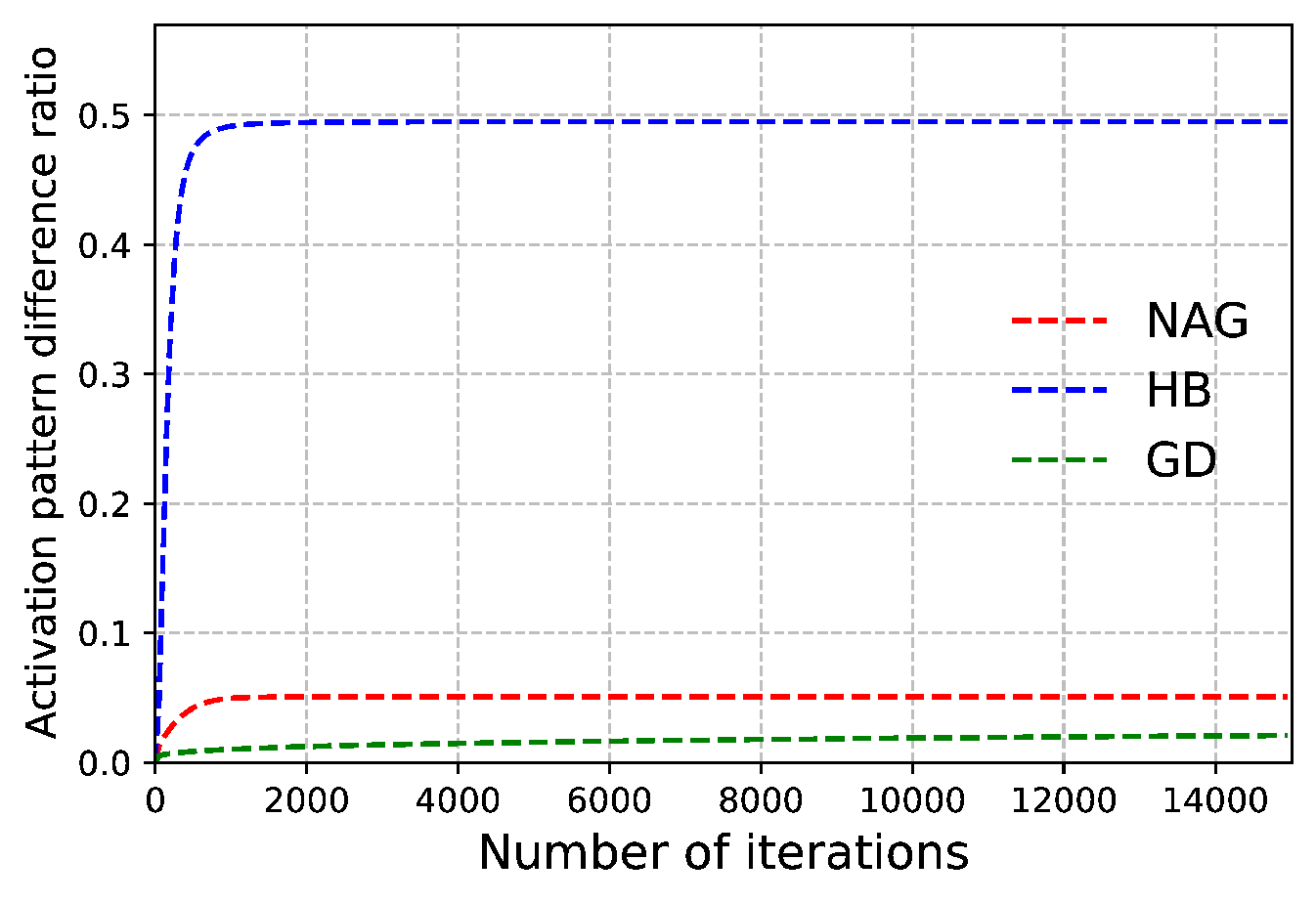}}\\
	\subfigure[CIFAR10]{\includegraphics[scale=0.47]{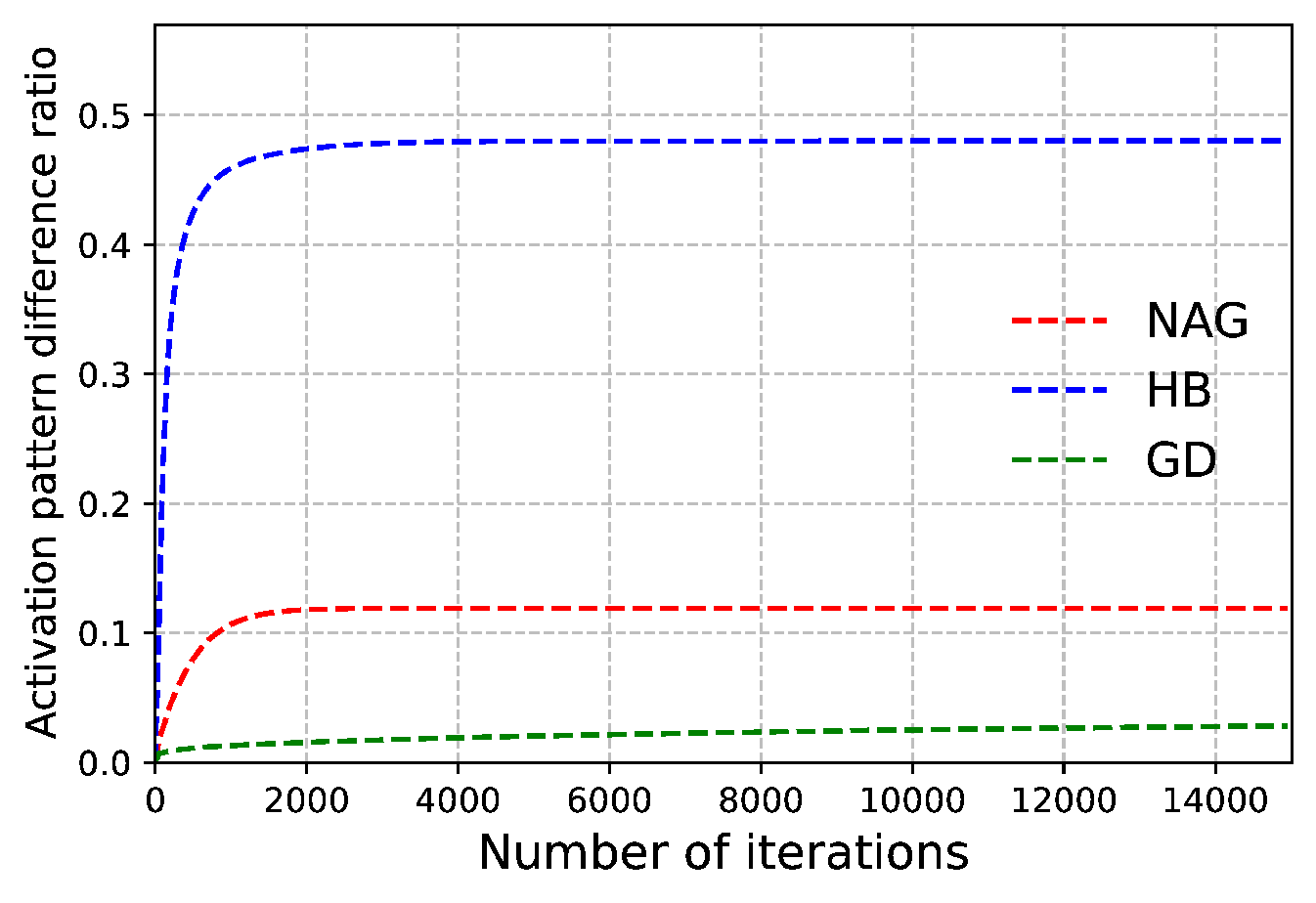}}
	\subfigure[ENERGY]{\includegraphics[scale=0.47]{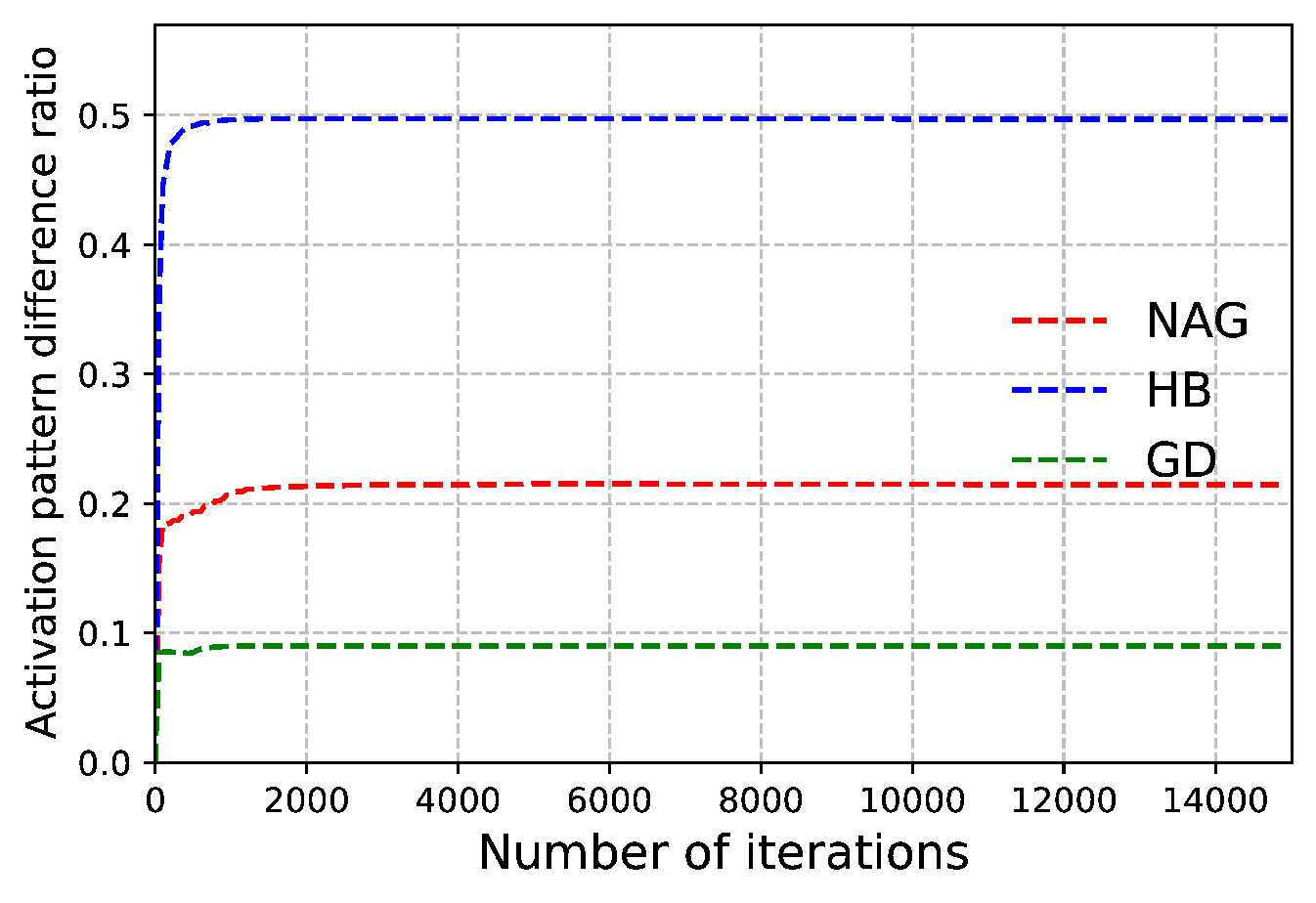}}\\
	\subfigure[HOUSING]{\includegraphics[scale=0.47]{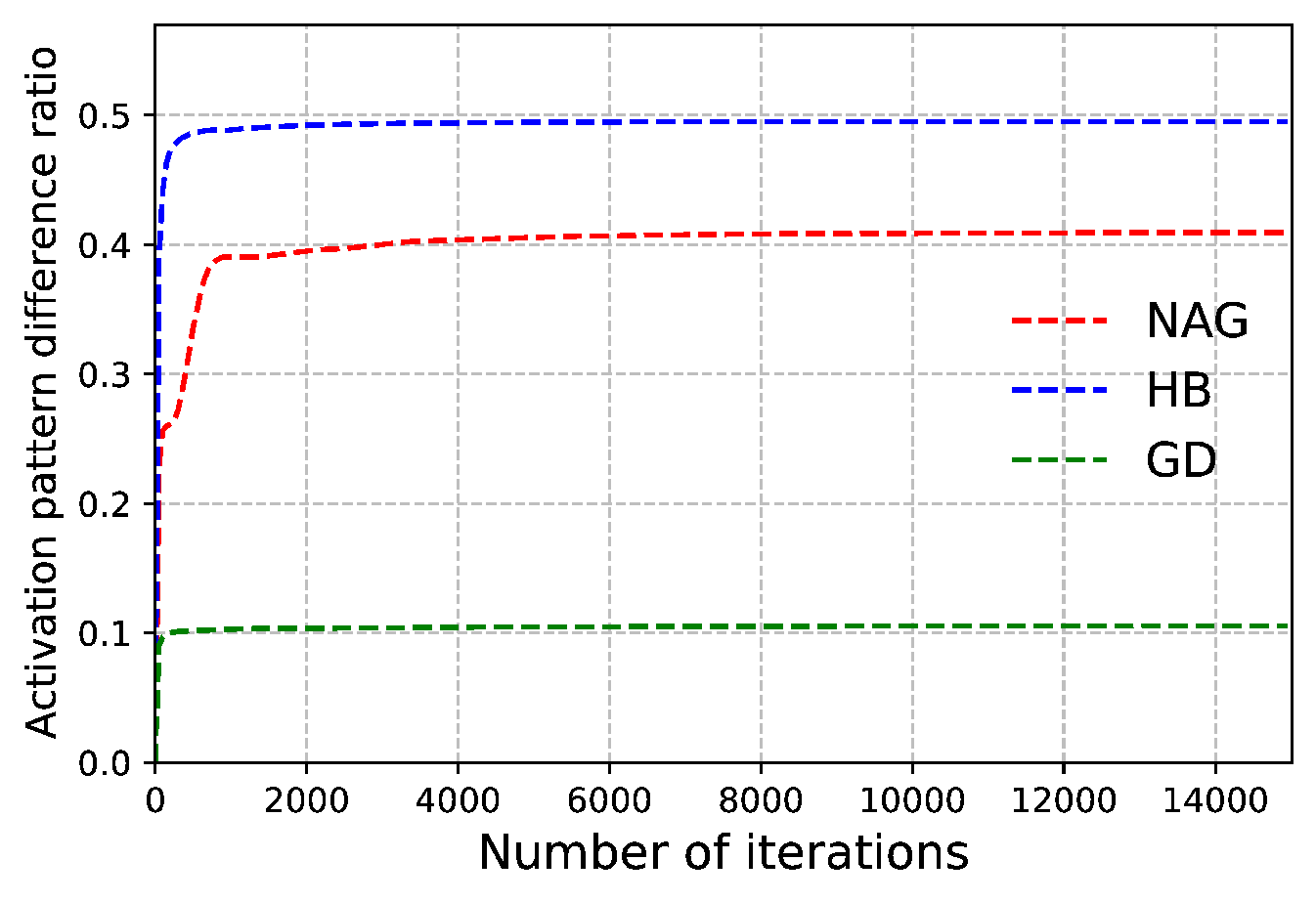}}
	\subfigure[YACHT]{\includegraphics[scale=0.47]{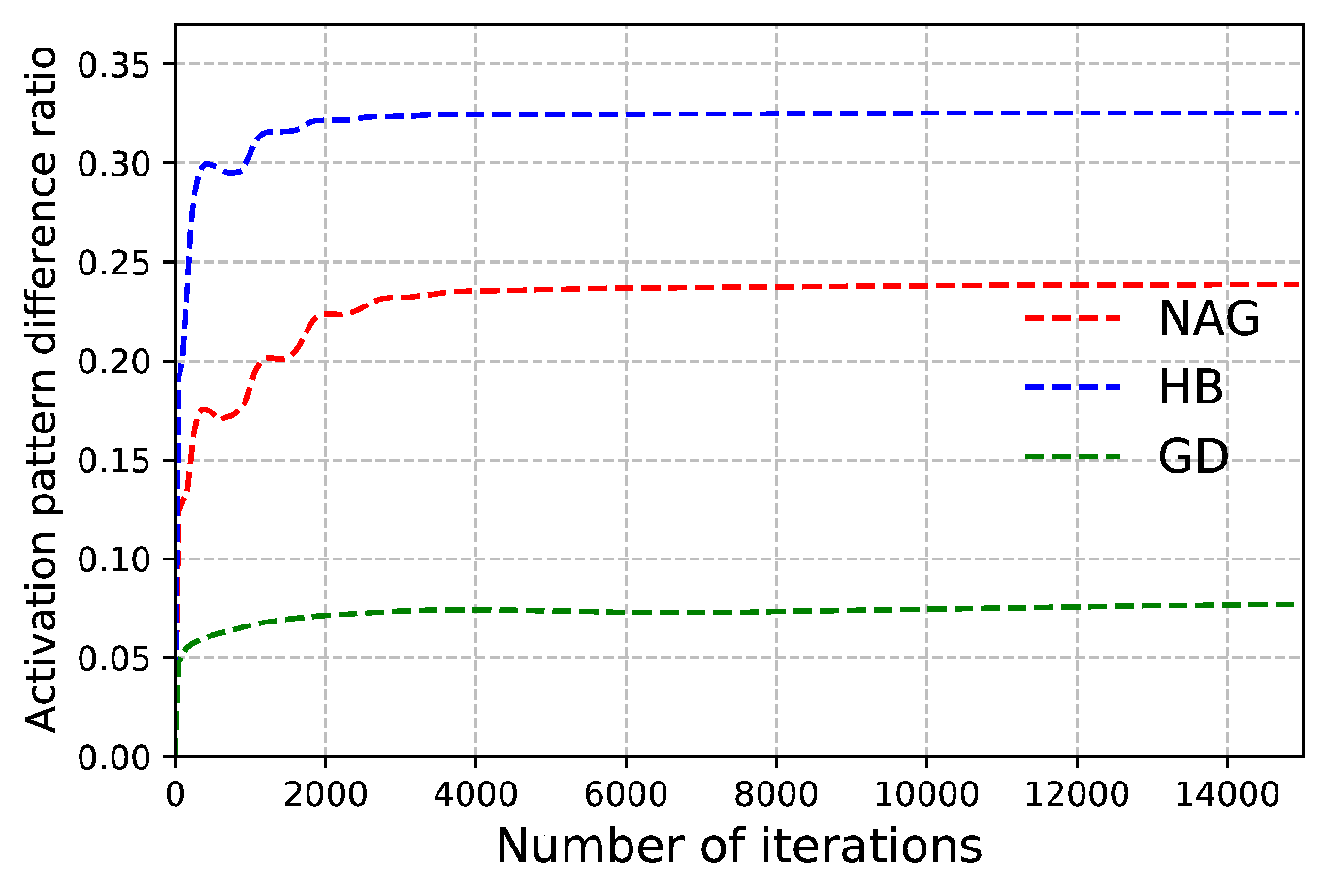}}
	\caption{Activation pattern difference ratio comparison among GD, HB and NAG with width $m=20000$.
	}
	\label{activation_pattern_d}
\end{figure*}

\subsection{Results analysis} 
\textbf{Convergence analysis}. We first provide the convergence comparison among NAG, HB and GD. 
The neural network is trained with 5 different initialization seed for each dataset, and the width of the hidden layer is $20000$.
The dashed line represents the mean training loss, while the shaded region represents the range of the maximum and minimum performance.
From Fig~\ref{Convergence}, we observe that NAG converges faster than GD on all six datasets.
Furthermore, it is noted that NAG achieves a comparable and even improved performance than HB.
This is in accordance with our theoretical findings.

\textbf{Impact of the over-parameterization}. Secondly, we evaluate the impact of the over-parameterization on two quantities relevant to our theoretical analysis.
One is the maximum distance $\max_{r \in [m]} \|\w_t^r - \w_0^r\|_2$, which is used to demonstrate the change of the parameter with respect to its initial~\cite{du2018gradient}.
The other is the activation pattern difference ratio $\frac{\sum_{i=1}^n\sum_{r=1}^m \mathbb{I}\{\mathbb{I}\{\langle \w_t^r, \x_i \rangle\} \neq \mathbb{I}\{\langle \w_0^r, \x_i \rangle\} \}}{mn}$.
It describes the percentiles of pattern changes among $mn$ patterns~\cite{du2018gradient,wang2020provable}.
In Remark 3, we theoretically show the the upper bounds of these two quantities are all scaled as $\mathcal{O}(1/\sqrt{m})$, indicating that the parameter stays closer to its initial as the width increases.
To observe the impact of the over-parameterization, we set the range of the width as $m \in [1250, 2500, 5000, 10000, 20000]$.
Each neural network of different width is trained with 5 different initialization seed, where the solid line indicates the corresponding mean value.
As shown in Fig~\ref{relative_distance} and Fig~\ref{activation_pattern}, the maximum distance and the activation pattern difference ratio both decrease as the width increases.

Moreover, we compare the above two quantities among NAG, HB and GD.
From Remark 3, we show that the upper bound of the maximum distance from the initialization for NAG is larger than that of GD by a factor $\mathcal{O}(\sqrt{\kappa})$, resulting in a larger activation pattern difference ratio for NAG over GD.
On the other hand, in Remark 3, we also show that NAG has comparable upper bounds for these two quantities as HB.
We conduct the experiments in the same setting as the convergence analysis.
According to Fig~\ref{relative_distance_d} and Fig~\ref{activation_pattern_d}, the two quantities of NAG are larger than that of GD.
Comparing to HB, NAG obtains a comparable or smaller values.
These phenomena support our theoretical results.

\section{Conclusion and future work}
In this paper, we focus on analyzing the training trajectory of NAG for optimizing a two-layer fully connected neural network with ReLU activation.
By exploiting the connection between the NTK and the finite over-parametrized neural network, we show that NAG can achieve a non-asymptotic linear convergence rate to a global optimum.
In the discrete-time scenario, our result provides theoretical guarantees for the acceleration of NAG over GD.
In addition, our result implies NAG obtains a comparable convergence rate as HB. 

An important future work is to extend our analysis to deep neural networks with different architectures (e.g., convolutional neural networks, graph neural networks, etc.) and activation functions (e.g., sigmoid, tanh, etc.).
Recently, there are plenty of works studied the convergence of GD on different types of over-paramterized neural networks~\cite{DBLP:conf/icml/DuLL0Z19, DBLP:conf/nips/DuHSPWX19,arora2019exact}.
The key technical challenge lies in deriving and analyzing the associated residual dynamics, which might be complex due to the structure of the neural network.  
Meanwhile, in practice, many applications requires numerous entities, yet their interactions are highly incomplete.
Currently, the latent factor model has attracted a lot of attention as a way to deal with this problem~\cite{9238448,9159907,9601264,9647958}.
It brings an interesting future direction for studying the acceleration of NAG for these problems, where the induced dynamics can be investigated using our approach.
\appendix

\section{Proof of Lemma \ref{lemma: rec form}}
\label{app:lemma_1}
\begin{proof}
At iteration $t+1$, the output of the two-layer fully connected neural network for arbitrary feature $\x_i$ can be divided into two parts w.r.t the set $S_i$
	\begin{eqnarray}
		&&f(\W_{t+1},\a;\x_i) \nonumber\\
	&=&\frac{1}{\sqrt{m}}\sum_{r=1}^m a^r \sigma (\langle	\w_{t+1}^{r}, \x_i \rangle) \nonumber \\
	\label{eq:twolayer}
		&=& \!\!\!\frac{1}{\sqrt{m}}\sum_{r \in S_i} a^r \sigma(\langle	\w_{t+1}^{r}, \x_i \rangle) \!+\! \frac{1}{\sqrt{m}}\sum_{r \in S_i^{\perp}} a^r \sigma(\langle	\w_{t+1}^{r}, \x_i \rangle). 
	\end{eqnarray}
For brevity, we define $\mathbb{I}_{r,i}(t)=\mathbb{I}\{{\langle	\w_t^r, \x_i \rangle \!\geq\! 0}\}$ and use $\frac{\partial \L}{\partial \w_t^r}$ instead of $\frac{\partial \L(\W_t, \a)}{\partial \w_t^r}$.
Based on the updating scheme (\ref{eq:NAG-SC_one_line}), the first part of ~(\ref{eq:twolayer}) can be decomposed as:
{{\begin{eqnarray}
&&\frac{1}{\sqrt{m}}\sum_{r \in S_i} a^r \sigma(\langle	\w_{t+1}^{r}, \x_i \rangle) \nonumber\\ 
&{=}& \frac{1}{\sqrt{m}}\sum_{r \in S_i} a^r\sigma(\langle	\w_t^{r} - \eta \frac{\partial \L}{\partial \w_t^{r}} + \beta(\w_t^{r} -\w_{t-1}^{r}) - \beta \eta (\frac{\partial \L}{\partial \w_t^{r}}- \frac{\partial \L}{\partial \w_{t-1}^{r}}), \x_i \rangle) \nonumber\\
&\overset{\text{a}}{=}& \frac{1}{\sqrt{m}}\!\sum_{r \in S_i}\! a^r\langle	\w_t^{r} - \eta \frac{\partial \L}{\partial \w_t^{r}} + \beta(\w_t^{r} -\w_{t-1}^{r}) - \beta \eta (\frac{\partial \L}{\partial \w_t^{r}}- \frac{\partial \L}{\partial \w_{t-1}^{r}}), \x_i \rangle \mathbb{I}_{r,i}(t+1)\nonumber\\
&\overset{\text{b}}{=}&\!\!\!\!\!\!{{\frac{1\!+\!\beta}{\sqrt{m}}\!\!\sum_{r \in S_i}\!a^r\langle \w_t^{r} , \x_i\rangle \mathbb{I}_{r,i}(t) \!-\! \frac{\beta}{\sqrt{m}}\!\!\sum_{r \in S_i}\!a^r\langle \w_{t\!-\!1}^{r} , \x_i\rangle\mathbb{I}_{r,i}(t\!-\!1)}} \!+\! \frac{\eta\beta}{\sqrt{m}}\!\!\sum_{r \in S_i}\!\!a^r\!\langle \frac{\partial \L}{\partial \w_{t\!-\!1}^{r}} , \x_i\rangle \mathbb{I}_{r,i}(t\!\!-\!\!1)  \nonumber\\
&-&\!\!\!\!\!\! \frac{(1\!+\!\beta)\eta}{\sqrt{m}}\!\!\sum_{r \in S_i}\!\!a^r\langle \frac{\partial \L}{\partial \w_t^{r}} , \x_i\rangle \mathbb{I}_{r,i}(t) \nonumber\\
\label{eq:S_i}
&\overset{\text{c}}{=}&\!\!\!\!\!\! (1+\beta)f(\W_t,\a;\x_i) - \frac{1+\beta}{\sqrt{m}}\sum_{r \in S_i^{\perp}}a^r\langle \w_t^{r} , \x_i\rangle \mathbb{I}_{r,i}(t)-\beta f(\W_{t-1},\a;\x_i) \nonumber \\
\!\!\!&+&\!\!\!\!\! \frac{\beta}{\sqrt{m}}\!\sum_{r \in S_i^{\perp}}a^r\langle \w_{t-1}^{r} , \x_i\rangle \mathbb{I}_{r,i}(t-1) \!-\!(1+\beta)\eta \sum_{j=1}^n\xib_t[j]\H_t[i,j] \!+\! \beta\eta \sum_{j=1}^n\xib_{t-1}[j]\H_{t-1}[i,j] \nonumber\\
\label{eq: decompose S_i}
&+& \!\!\!\!\!\!\frac{(1+\beta)\eta}{m}\!\!\sum_{j=1}^n \x_i^{\top}\x_j\xib_t[j]\!\!\sum_{r \in S_i^{\perp}}\!\mathbb{I}_{r,i}(t)\mathbb{I}_{r,j}(t) \!-\! \frac{\beta\eta}{m}\!\!\sum_{j=1}^n \x_i^{\top}\x_j\xib_{t-1}[j]\!\!\sum_{r \in S_i^{\perp}}\!\mathbb{I}_{r,i}(t\!-\!1)\mathbb{I}_{r,i}(t\!-\!1),
\end{eqnarray} }}
where (a) applies the property of ReLU activation $\sigma(x) = x\mathbb{I}\{x\geq 0\}$,\\
(b) exploits the the neurons of the set $S_i$ always keep the sign of their activation pattern,\\
(c) uses the expansion of the subgradient~(\ref{eq:gradient_objective}) and the Gram matrix $\H_t$~(\ref{eq:gram matrix h_t}).

With (\ref{eq: decompose S_i}), the $i$-th element of the residual error  can be decomposed as: 
{{
\begin{eqnarray}
&&\xib_{t+1}[i] \nonumber\\ 
 &=&  f(\W_{t+1},\a;\x_i) - y_i \nonumber\\
&=&\!\!\!\! \frac{1}{\sqrt{m}}\!\!\sum_{r \in S_i}\!\! a^r \sigma(\langle	\w_{t+1}^{r}, \x_i \rangle)\!+\! \frac{1}{\sqrt{m}}\!\!\sum_{r \in S_i^{\perp}}\!\! a^r \sigma(\langle	\w_{t+1}^{r}, \x_i \rangle) \!-\!y_i \nonumber \\
&=& \xib_{t}[i] + \beta(\xib_{t}[i] - \xib_{t-1}[i])-(1+\beta)\eta\sum_{j=1}^n \H_0[i,j]\xib_t[j] + \beta\eta \!\!\sum_{j=1}^n \!\!\H_0[i,j]\xib_{t-1}[j] \nonumber \\
&+& \beta\eta \!\!\sum_{j=1}^n (\H_{t-1}[i,j]\!-\!\H_0[i,j])\xib_{t-1}[j] - (1+\beta)\eta \!\!\sum_{j=1}^n (\H_{t}[i,j]\!\!-\!\!\H_0[i,j])\xib_{t}[j] \nonumber \\
&+&\!\!\frac{(1\!\!+\!\!\beta)\eta}{m}\!\!\sum_{j=1}^n\!\! \x_i^{\top}\x_j\xib_t[j]\!\!\sum_{r \in S_i^{\perp}}\!\!\mathbb{I}_{r,i}(t)\mathbb{I}_{r,j}(t) - \frac{\beta\eta}{m}\!\!\sum_{j=1}^n \!\!\x_i^{\top}\x_j\xib_{t-1}[j]\!\!\sum_{r \in S_i^{\perp}}\!\!\mathbb{I}_{r,i}(t-1)\mathbb{I}_{r,j}(t-1) \nonumber \\
&+&\!\!\!\! \frac{1}{\sqrt{m}}\!\!\!\sum_{r \in S_i^{\perp}}\!\! a^r \sigma(\langle	\w_{t+1}^{r}, \x_i \rangle)- a^r\sigma(\langle \w_t^{r} , \x_i\rangle)-\frac{\beta}{\sqrt{m}}\!\!\sum_{r \in S_i^{\perp}}\!\!a^r\sigma(\langle \w_t^{r} , \x_i\rangle ) - a^r\sigma(\langle \w_{t-1}^{r} , \x_i\rangle )\nonumber
\end{eqnarray}}}
In words, the residual error on the whole training dataset can be written in a recursive form as:
\begin{eqnarray}
	\xib_{t+1} &=& \xib_t + \beta(\xib_t - \xib_{t-1}) - (1+\beta)\eta \H_0\xib_t + \beta\eta \H_{0}\xib_{t-1}  + \psib_t + \phib_t\nonumber
\end{eqnarray}
where $\psib_t = \beta\eta(\H_{t-1} - \H_0)\xib_{t-1} -(1+\beta)\eta(\H_t - \H_0)\xib_t$ and the i-th element of $\phib_t$ have the following form
\begin{eqnarray}
	\phib_t[i] \!\!\!\!&=& \!\!\!\! \frac{(1\!\!+\!\!\beta)\eta}{m}\!\!\sum_{j=1}^n\!\! \x_i^{\top}\x_j\xib_t[j]\!\!\sum_{r \in S_i^{\perp}}\!\!\mathbb{I}_{r,i}(t)\mathbb{I}_{r,j}(t) - \frac{\beta\eta}{m}\!\!\sum_{j=1}^n\!\! \x_i^{\top}\x_j\xib_{t-1}[j]\!\!\!\sum_{r \in S_i^{\perp}}\!\!\mathbb{I}_{r,i}(t-1)\mathbb{I}_{r,j}(t-1) \nonumber \\
&+& \!\!\!\!\frac{1}{\sqrt{m}}\!\!\sum_{r \in S_i^{\perp}}\!\! a^r \sigma(\langle	\w_{t+1}^{r}, \x_i \rangle)\!-\! a^r\sigma(\langle \w_t^{r} , \x_i\rangle) \!-\! \frac{\beta}{\sqrt{m}}\!\!\sum_{r \in S_i^{\perp}}\!\!a^r\sigma(\langle \w_t^{r} , \x_i\rangle ) \!-\! a^r\sigma(\langle \w_{t-1}^{r} , \x_i\rangle ).\nonumber 
\end{eqnarray}

In addition, it has $\sum_{r \in S_i^{\perp}}\mathbb{I}_{r,i}(t)\mathbb{I}_{r,j}(t) \leq |S_i^{\perp}|$.
Using $\|\x_i\| \leq 1$ and 1-Lipschitz property of ReLU activation $\sigma(\cdot)$, it has
\begin{eqnarray}
 a^r\big[\sigma(\langle	\w_{t+1}^{r}, \x_i \rangle)- \sigma(\langle \w_t^{r} , \x_i\rangle )\big]  &\leq& |a^r \big[\sigma(\langle	\w_{t+1}^{r}, \x_i \rangle )-\sigma(\langle \w_t^{r} , \x_i\rangle)\big]| \nonumber \\
&\leq&|\langle \w_{t+1}^r - \w_t^r, \x_i \rangle| \nonumber\\
&\leq&\|\w_{t+1}^r - \w_t^r\|\|\x_i\| \nonumber\\
&\leq&\|\w_{t+1}^r - \w_t^r\| \nonumber.
\end{eqnarray}

Therefore, we have
\begin{eqnarray}
&&| \phib_t[i] | \nonumber \\
&{\leq}&\!\!\!\! \frac{(1+\beta)\eta|S_i^{\perp}|}{m} \sum_{j=1}^n |\xib_t[j]| \!+\! \frac{\beta\eta |S_i^{\perp}|}{m}  \sum_{j=1}^n |\xib_{t-1}[j]| \!+\! \frac{1}{\sqrt{m}}\!\! \sum_{r \in S_i^{\perp}}( \| \w_{t+1}^{r} \!-\! \w_t^r \| \!+\! \beta \|\w_t^{r} \!-\!  \w_{t-1}^r\|\!)\nonumber \\
&\overset{\text{a}}{\leq}&\!\!\!\! \frac{(1+\beta)\sqrt{n}\eta|S_i^{\perp}|}{m}\|\xib_t\|\!\!+\!\!\frac{\sqrt{n}\beta \eta|S_i^{\perp}|}{m}\|\xib_{t-1}\| + \frac{1}{\sqrt{m}} \sum_{r \in S_i^{\perp}}( \| \w_{t+1}^{r} - \w_t^{r} \|+ \beta\|\w_t^{r} - \w_{t-1}^{r}\|) \nonumber \\
\label{lemma_1:phi_t}
&\overset{\text{b}}{\leq}&\!\!\! \frac{ \sup\limits_{j\in [n]}|S_j^{\perp}|\sqrt{n}\eta}{m}((2+4\beta)\|\xib_t\|+ 2\beta\|\xib_{t-1} +2\sum_{i=0}^{t-1}\beta^{t+1-i}\|\xib_i\|)\nonumber 
\end{eqnarray}
where 
(a) uses $\sum_{j=1}^n |z_j| \leq \sqrt{n\sum_{j=1}^n z_j^2}$,\\
(b) uses the NAG update rule to derive the distance between two consecutive iterations
\begin{eqnarray}
\w_{t}^{r} - \w_{t-1}^{r} &=& -\eta \frac{\partial \L}{\partial \w_{t-1}^{r}} + \beta(\v_{t}^{r} - \v_{t-1}^{r})  \nonumber \\
	&=& -\eta \frac{\partial \L}{\partial \w_{t-1}^{r}} + \beta\big(-\eta\frac{\partial \L}{\partial \w_{t-1}^{r}}+ \beta(\v_{t-1}^{r} - \v_{t-2}^{r})\big) \nonumber\\
	&=& -\eta (1+\beta)\frac{\partial \L}{\partial \w_{t-1}^{r}} + \beta^2(\v_{t-1}^{r} - \v_{t-2}^{r})\nonumber\\
	&=& -\eta(1+\beta)\frac{\partial \L}{\partial \w_{t-1}^{r}}+\beta^2\big(-\eta\frac{\partial \L}{\partial \w_{t-2}^{r}} +  \beta(\v_{t-2}^{r} - \v_{t-3}^{r})\big) \nonumber\\
	\label{NAG:two consecutive}
&=&-\eta\frac{\partial \L}{\partial \w_{t-1}^{r}}   -\eta\sum_{i=0}^{t-1}\beta^{t-i}\frac{\partial \L}{\partial \w_{i}^{r}}. 
\end{eqnarray}
Therefore,
\begin{eqnarray}
\|\w_{t}^{r} - \w_{t-1}^{r}\|	\!\!\!&\leq&\!\!\! \eta \|\frac{\partial \L}{\partial \w_{t-1}^{r}}\| + \eta\sum_{i=0}^{t-1}\beta^{t-i}\|\frac{\partial \L}{\partial \w_{i}^{r}}\| \leq \eta \frac{\sqrt{n}}{\sqrt{m}}(\|\xib_t\| + \sum_{i=0}^{t-1}\beta^{t-i}\|\xib_i\| ),
\end{eqnarray}
where $|\frac{\partial \L}{\partial \w_{i}^{r}}| \leq \frac{\sqrt{n}}{\sqrt{m}}\|\xib_i\|$ and $\v_1^{r} - \v_0^{r}=-\eta \frac{\partial \L}{\partial \w_{0}^{r}}$.
Note that 
\begin{eqnarray}
	\w_{t+1}^{r} - \w_t^{r} &=& \beta(\w_{t}^{r} - \w_{t-1}^{r}) - \eta(1+\beta)\frac{\partial \L}{\partial \w_{t}^{r}} + \eta\beta\frac{\partial \L}{\partial \w_{t-1}^{r}},\nonumber
\end{eqnarray}
then we have
\begin{eqnarray}
	 \| \w_{t+1}^{r} - \w_t^{r} \|+ \beta\|\w_t^{r} - \w_{t-1}^{r}\| 
	\!\! &\leq&\!\!\!\!  \eta(1\!\!+\!\!\beta)\|\frac{\partial \L}{\partial \w_{t}^{r}}\| + \eta\beta\|\frac{\partial \L}{\partial \w_{t-1}^{r}}\| \!+\! 2\beta\|\w_t^{r} \!\!-\!\! \w_{t-1}^{r}\|   \nonumber\\
	 &\leq&\!\!\!\! \frac{\eta\sqrt{n}}{\sqrt{m}}((1+3\beta)\|\xib_t\| + \beta\|\xib_{t-1} \| + \!2\sum_{i=0}^{t-1}\beta^{t+1-i}\|\xib_i\|).\nonumber
\end{eqnarray}
\end{proof}

\section{Proof of Lemma \ref{lemma: matrix_vector}}
\label{app: matrix_vector}
\begin{proof}
First,  
we decompose $\H = \U \Lambda \U^*$ with SVD method, where $\U$ is an unitary matrix and $\Lambda = diag(\lambda_1,\cdots, \lambda_n)$ is a diagonal matrix, $\lambda_i$ is the i-th eigenvalues of $\H$ in a decreasing order.
Then we have
{{
\begin{eqnarray}
\M = 
\begin{bmatrix}
   \U & \textbf{0}_n \\
   \textbf{0}_n & \U 
   \end{bmatrix}
\begin{bmatrix}
   (1+\beta)(\I_n-\eta \Lambda) &
  \beta(-\I_n+\eta \Lambda) \\
   \I_n & \textbf{0}_n 
   \end{bmatrix}
   \begin{bmatrix}
   \U^* & \textbf{0}_n \\
   \textbf{0}_n & \U^*   \end{bmatrix}.\nonumber
\end{eqnarray}
We define $\tilde{\U} =$ {\small{$\begin{bmatrix}
   \U & \textbf{0}_n \\
   \textbf{0}_n & \U 
   \end{bmatrix}$}}}}.
By applying some permutation matrix $\tilde{\P}$, M can be further decomposed as:
\begin{eqnarray}
	\M = \tilde{\U}\tilde{\P}\Sigma\tilde{\P}^{\top}\tilde{\U}^*,
\end{eqnarray}
where $\mathbf{\Sigma}$ is a block diagonal matrix with ${\mathbf{\Sigma}}_i =$ {\small{$\begin{bmatrix}
   (1+\beta)(1-\eta\lambda_i) & \beta(-1+\eta\lambda_i) \\
   1 & 0 
   \end{bmatrix}$}}.
   
After applying eigendecomposition method, $\mathbf{\Sigma}_i$ can be factorized as:
\begin{eqnarray}
	{\mathbf{\Sigma}}_i = \Q_i \D_i \Q_i^{-1},
\end{eqnarray}
where the columns of $\Q_i$ are the eigenvectors of ${\mathbf{\Sigma}}_i$ and $\D_i$ is a diagonal matrix whose diagonal elements are the corresponding eigenvalues.
Then, $\mathbf{\Sigma}$ can be written as:
\begin{eqnarray}
\mathbf{\Sigma} &=& diag({\mathbf{\Sigma}}_1, \cdots, {\mathbf{\Sigma}}_n)=\Q \D\Q^{-1}
\end{eqnarray}
where $\Q=diag(\Q_1, \cdots, \Q_n)$ and $\D=diag(\D_1, \cdots, \D_n)$.  
As a result, we have
\begin{equation}
\label{eq: decompose of M}
	\M = \P\D\P^{-1}, 
\end{equation}
where $\P=\tilde{\U}\tilde{\P}\Q$.

Now, we provide the bound for the norm of $\v_k$. 
We define $\u_k = \P^{-1}\v_k$.
Substituting the expression of $\v_k$  and (\ref{eq: decompose of M}) into  $\u_k$, we have
\begin{eqnarray}
\label{eq: defi u_k}
\u_k = \P^{-1}\M\v_{k-1}=\D \u_{k-1}=\D^k \u_{0}. 
\end{eqnarray}
Plugging the definition of $\u$ back into (\ref{eq: defi u_k}), we have 
\begin{eqnarray}
	\P^{-1}\v_k &=& \D^k \P^{-1}\v_{0} \nonumber \\
	\v_k &=& \P \D^k \P^{-1} \v_{0} \nonumber \\
	\label{eq:v_k}
 \|\v_k\| &\leq& (\max_{i \in [n]} |\D_{ii}|^k)\sqrt{\frac{\lambda_{max}(\P\P^*)}{\lambda_{min}(\P\P^*)}}\|\v_0\|.
\end{eqnarray}
Note that the right-hand side of~(\ref{eq:v_k}) is determined by the $\max_{i \in [n]}|\D_{ii}|$, the condition number of $\P\P^*$ and $\|\v_0\|$.\\

Next, we analyze the choice of $\eta$ and $\beta$ that guarantees $\max_{i \in [n]}|\D_{ii}| < 1$.
Note that the characteristic polynomial of $\mathbf{\Sigma}_i$ is $z^2 -(1+\beta)(1-\eta\lambda_i)z + \beta(1-\eta\lambda_i)$.
If $\Delta_i=((1+\beta)(1-\eta \lambda_i))^2 - 4\beta(1-\eta \lambda_i) \leq 0$, the two roots $z_{i,1}$ and $z_{i,2}$ are conjugate with the same magnitude $\sqrt{\beta(1-\eta\lambda_i)}$.
According to the sign of $\Delta_i$, it is easy to show
\begin{eqnarray}
\Delta_i \leq 0 &\Leftrightarrow& (1+\beta)^2(1-\eta\lambda_i-\frac{2\beta}{(1+\beta)^2})^2 - (1+\beta)^2(\frac{2\beta}{(1+\beta)^2})^2 \leq 0 \nonumber \\
   \label{eq:eta}
   &\Leftrightarrow& 0 \leq 1-\eta\lambda_i \leq \frac{4\beta}{(1+\beta)^2}.       
\end{eqnarray}
For all $i \in [n]$, we have the constraints on $\eta$ and $\beta$ as:
\begin{eqnarray}
\label{bound:beta}
	0<\eta \leq 1/\lambda_{max}(\H),\;\;  1 \geq \beta \geq \frac{1-\sqrt{\eta\lambda_{min}(\H)}}{1+\sqrt{\eta\lambda_{min}(\H)}}.
\end{eqnarray}
Then we have $\max_{i \in [n]}|\D_{ii}| = \sqrt{\beta(1-\eta\lambda_{min}(\H))}< 1$.

Next, we provide the bounds of the eigenvalues for $\P\P^*$.
Note that the spectrum of Q does not change by multiplying the unitary matrix $\tilde{\U}\tilde{\P}$.
Therefore, we turn to analyze the eigenvalues of $\Q\Q^*$ instead of $\P\P^*$.
We define $\lambda_{max}(\Q\Q^*) = \max_{i \in [n]} \lambda_{max}(\Q_i\Q_i^*)$ and $\lambda_{min}(\Q\Q^*) = \min_{i \in [n]} \lambda_{min}(\Q_i\Q_i^*)$.
The two eigenvalues $z_{i,1}$ and $z_{i,2}$ of $\mathbf{\Sigma}_i$ satisfy
\begin{eqnarray}
	z_{i,1} + z_{i,2} = (1+\beta)(1-\eta\lambda_i),\;\;
	z_{i,1}  z_{i,2} = \beta(1-\eta\lambda_i). \nonumber
\end{eqnarray}
For eigenvalue $z_{i,j}$, the corresponding eigenvector is  $q_{i,j} = [z_{i,j}, 1]^{\top}$.
As a result, we have
\begin{eqnarray}
\Q_i\Q_i^* \!=\! q_{i,1}q_{i,1}^{*} + q_{i,2}q_{i,2}^{*} \!=\! {{\begin{bmatrix}
   z_{i,1}\bar{z}_{i,1}+z_{i,2}\bar{z}_{i,2} & z_{i,1} + z_{i,2}
 \\
   \bar{z}_{i,1} + \bar{z}_{i,2} & 2 
   \end{bmatrix}}} . \nonumber
\end{eqnarray}
We denote $\theta_{i,1}$ and $\theta_{i,2}$ as the two eigenvalues of $\Q_i\Q_i^*$, then
\begin{eqnarray}
	\theta_{i,1} + \theta_{i,2} &=&  z_{i,1}\bar{z}_{i,1}+z_{i,2}\bar{z}_{i,2}+2 \nonumber\\
	\theta_{i,1} \theta_{i,2} &=& 2( z_{i,1}\bar{z}_{i,1}+z_{i,2}\bar{z}_{i,2}) - (z_{i,1} + z_{i,2})( \bar{z}_{i,1} + \bar{z}_{i,2}) \nonumber 
\end{eqnarray}
The matrix $\Q_i\Q_i^*$ is positive semi-definite, we have
\begin{equation}
\theta_{i,1} + \theta_{i,2} \geq \max\{\theta_{i,1}, \theta_{i,2}\} \geq \frac{\theta_{i,1} + \theta_{i,2}}{2}.
\end{equation}
Moreover, one can get the minimum of the two eigenvalues as
\begin{eqnarray}
	\min\{\theta_{i,1}, \theta_{i,2}\}= \theta_{i,1}\theta_{i,2}/\max\{\theta_{i,1}, \theta_{i,2}\} .
\end{eqnarray}
For bounding the eigenvalues of $QQ^*$, we have
\begin{eqnarray}
	\lambda_{max}(QQ^*) \leq \max_{i \in [n]}\{\max\{\theta_{i,1}, \theta_{i,2}\}\} 
	&\leq& \max_{i \in [n]}\{\theta_{i,1} + \theta_{i,2}\} 
	\leq 2\beta(1-\eta\lambda_{min}(\H)) + 2, \nonumber
\end{eqnarray}
and
\begin{eqnarray}
	\lambda_{min}({QQ^*}) \geq \min_{i \in [n]}\{\min\{\theta_{i,1}, \theta_{i,2}\}\} 
	&\geq& \min_{i \in [n]}\{\theta_{i,1} \theta_{i,2}/\max\{\theta_{i,1}, \theta_{i,2}\}\} \nonumber \\
	&\geq& \frac{\min_{i \in [n]}\{\theta_{i,1} \theta_{i,2}\}}{\max_{i \in [n]}\{\theta_{i,1},\theta_{i,2}\}} \nonumber\\
	&\geq& \frac{\min_{i \in [n]}\{4\beta(1-\eta\lambda_i) - [(1+\beta)(1-\eta\lambda_i)]^2\}}{2\beta(1-\eta\lambda_{min}(\H)) + 2} \nonumber \\
	&\geq& \frac{\min\{g(\beta, \eta\lambda_{min}(\H)), g(\beta, \eta\lambda_{max}(\H))\}}{2\beta(1-\eta\lambda_{min}(\H)) + 2} ,\nonumber
\end{eqnarray}
where the last inequality applies $g$ is a concave quadratic function with respect to $1- \eta\lambda_i$ when $\beta >= 0$ and the minimum value must be found at the boundary.
Therefore, 
\begin{eqnarray}
	\frac{\lambda_{max}(QQ^*)}{\lambda_{min}({QQ^*})} \leq \frac{(2\beta(1-\eta\lambda_{min}(\H)) + 2)^2}{\min\{g(\beta, \eta\lambda_{min}(\H)), g(\beta, \eta\lambda_{max}(\H))\}},\nonumber
\end{eqnarray}
which completes the proof.
\end{proof}

\section{Proof of Lemma \ref{lemma: specific setting}}
\label{app: specific setting}

\begin{proof}
	With $\eta = 1/(2\lambda_{max})$ and $\beta = \frac{3\sqrt{{\kappa}} - 2}{3\sqrt{{\kappa}} + 2}$, we have
\begin{eqnarray}
	\sqrt{\beta(1-\eta\lambda_{min}(\H))} \leq \sqrt{\beta(1-\eta\lambda)} 
	&\leq& \sqrt{\frac{3\sqrt{{\kappa}} - 2}{3\sqrt{{\kappa}} + 2} (1 - \frac{1}{2{\kappa}})} \nonumber \\
	&\leq& \sqrt{\frac{3\sqrt{{\kappa}} - 2}{3\sqrt{{\kappa}} + 2} \frac{9{\kappa} - 4}{9{\kappa}}} \nonumber\\
	&=& 1 - \frac{2}{3\sqrt{{\kappa}}}, \nonumber
\end{eqnarray}
where the last inequality uses $1 - 1/(2{\kappa}) \leq (9{\kappa} - 4)/(9{\kappa})$.

Recall  $ C  = \frac{2\beta(1-\eta\lambda_{min}(\H)) + 2}{\sqrt{\min\{g(\beta, \eta\lambda_{min}(\H)), g(\beta, \eta\lambda_{max}(\H))\}}}$ and the function $g$ is defined as $g(x, y) = 4x(1-y) - [(1+x)(1-y)]^2$.
We have 
{{
\begin{eqnarray}
	1-\eta\lambda\geq 1-\eta\lambda_{min}(\H) \geq 1-\eta\lambda_{max}(\H) \geq1 - \eta\lambda_{max} \geq \frac{2\beta}{(1+\beta)^2}. \nonumber
\end{eqnarray}}}

Since $g$ is a concave quadratic function, then we have 
\begin{eqnarray}
	\min\{g(\beta, \eta\lambda_{min}(\H)), g(\beta, \eta\lambda_{max}(\H))\} &\geq& g(\beta, \eta\lambda) \geq\frac{2{\kappa} - 1}{{\kappa} (3\sqrt{{\kappa}} + 2)^2}. \nonumber
\end{eqnarray}
As a result, we have the upper bound
\begin{eqnarray}
	C &\leq& \frac{2 \frac{3\sqrt{{\kappa}} -2}{3\sqrt{{\kappa}} +2} \frac{2{\kappa} - 1}{2{\kappa}} + 2}{\sqrt{\frac{2{\kappa} - 1}{{\kappa} (3\sqrt{{\kappa}} + 2)^2}}} \leq \frac{2(9{\kappa}\sqrt{{\kappa}} - 3\sqrt{\bar{\kappa}} - 2{\kappa} + 2)}{\sqrt{{\kappa}(2{\kappa} - 1)}} \leq 12\sqrt{{\kappa}},
\end{eqnarray}
which completes the proof.
\end{proof}

\section{Supporting Lemmas}
\label{section: supporting lemmas}

\begin{lemma}(Claim 3.12 in \cite{song2019quadratic})
\label{lemma: bound of S_i}
 Suppose for all $t \in [T]$ and $r \in [m]$, $\|\w_t^{r} - \w_0^{r}\| \leq R$, where $R \in (0, 1)$. 
With probability at least $1-n \cdot exp(-mR)$, we have the following bound for all $i \in [n]$ as:
\begin{equation}
|S_i^{\perp}| \leq 4mR.
\end{equation}
\end{lemma}

\begin{lemma}(Lemma 3.2 in \cite{song2019quadratic})
\label{lemma: H_t and H_0}
Assume $\w_0^{r} \sim N(0, \I_d)$ for all $r \in [m]$.
Suppose for any set $\W = \{\w^1, \cdots, \w^m\}$ that satisfy $\|\w^r - \w_0^r\| \leq R$ for all $r \in [m]$, then it has
\begin{eqnarray}
	\|\H - \H_0\|_F \leq 2nR
\end{eqnarray}
with probability at least $1-n^2 \exp (-mR/10)$.
\end{lemma}

\begin{lemma}(Claim 3.10 in \cite{song2019quadratic}) 
\label{lemma: init bound}
Assume that $\w_0^{r} \sim N(0, \I_d)$ and $a^r$ is uniformly sampled form $\{-1,1\}$ for all $r \in [m]$. 
For $0 < \delta < 1$, we have
\begin{equation}
	\|\xib_0\|^2 = \mathcal{O}(nlog(m/\delta)log^2(n/\delta))
\end{equation}
with probability at least $1-\delta$.
\end{lemma}

\section{Proof of Theorem 1}

\begin{proof}
For simplicity, we define $ \alpha = 1 - \frac{2}{3\sqrt{\kappa}}$
and $\rho = \alpha + \frac{1}{6\sqrt{\kappa}} = 1 - \frac{1}{2\sqrt{\kappa}}$.
Then we have $\beta < \rho^2$.

Our goal is to prove the residual error dynamics follows a linear convergence form as:
\begin{eqnarray}
\label{z}
\|\z_{s}\| \leq \rho^s 2\gamma\|\z_0\|,
\end{eqnarray}
where $\z_{s} = [
   \xib_{s} ;
   \xib_{s-1}] 
$, $0 < \rho < 1$ and $\gamma$ is a positive constant.
We will prove the (\ref{z}) by induction.

The base case when $s = 0$  trivially holds.
For the induction step,  now we assume $\|z_s\| \leq \rho^s 2\gamma\|z_0\|$ for any time $ s \leq t-1$.
We argue it also holds at time t.

From Lemma 1, at iteration t, we have:
\begin{eqnarray}
\z_{t} &=& \M\z_{t-1} + \mub_{t-1}=\M^t\z_{0} + \sum_{s=0}^{t-1}\M^{t-s-1}\mub_{s} \nonumber \\
\label{bound:z_t}
\|\z_{t}\| &\leq& \|\M^t\z_{0}\| + \|\sum_{s=0}^{t-1}\M^{t-s-1}\mub_{s}\|. 
\end{eqnarray}
The next step is to separately analyze the two parts of the right-hand side of~(\ref{bound:z_t}).

By applying Lemma~\ref{lemma: matrix_vector}, we have the bound of $\|\M^t\z_0\|$ as:
\begin{eqnarray}
\label{bound: M_z}
	\|\M^t \z_0\| \leq \alpha^t \gamma \|\z_0\| \leq \rho^t \gamma \|\z_0\|.
\end{eqnarray}

Then we turn to provide an upper bound for  $\|\sum_{s=0}^{t-1}\M^{t-s-1}\mub_{s}\|$.
Before proving that, we first bound the distance between $\w_s^r$ and the initial $\w_0^r$ for all $s \leq t$ and $r \in [m]$.
Based on~(\ref{NAG:two consecutive}), we have
\begin{eqnarray}
	\w_{s}^{r} - \w_0^{r}  &=& \sum_{i=1}^s (\w_{i}^{r} - \w_{i-1}^{r}) = -\eta\!\sum_{i=0}^{s-1}\frac{\partial \L}{\partial \w_{i}^{r}} -\eta\!\sum_{g=0}^{s-1} \sum_{i=0}^g \beta^{g+1-i}\frac{\partial \L}{\partial \w_{i}^{r}}. \nonumber 
\end{eqnarray}
Applying Cauchy-Schwarz inequality and $|\frac{\partial \L}{\partial \w_{i}^{r}}| \leq \frac{\sqrt{n}}{\sqrt{m}}\|\xib_i\|$ , we have the bound of the distance for all $s \leq t$ and $r \in [m]$
\begin{eqnarray}
	\|\w_{s}^{r} - \w_0^{r}\|
	 &\leq& \frac{\eta\sqrt{n}}{\sqrt{m}}\sum_{i=0}^{s-1} \|\xib_i\|+ \frac{\eta\sqrt{n}}{\sqrt{m}}\sum_{g=0}^{s-1}\sum_{i=0}^g \beta^{g+1-i}\|\xib_i\| \nonumber\\
	&\overset{\text{a}}{\leq}&\frac{2\gamma\eta\sqrt{2n}}{\sqrt{m}}\|\xib_0\|\big( \sum_{i=0}^{s-1}\rho^i +\sum_{g=0}^{s-1}\sum_{i=0}^g \beta^{g+1-i}\rho^i \big) \nonumber \\
	&\overset{\text{b}}{\leq}& \frac{2\gamma\eta\sqrt{2n}}{\sqrt{m}}\|\xib_0\|\big( \frac{1}{1- \rho} + \frac{\rho^2}{(1-\rho)^2} \big)  \nonumber \\
	&\overset{\text{c}}{\leq}& \frac{8\kappa\gamma\eta\sqrt{2n}}{\sqrt{m}}\|\xib_0\| \nonumber \\	
	&=& \frac{48\sqrt{2n\kappa}}{\lambda\sqrt{m}}\|\xib_0\|
\end{eqnarray}
where
(a) uses the inductive hypothesis for $s \leq t-1$,\\
(b) uses $\beta \leq \rho^2$,\\
(c) applies $\frac{1}{1-\rho} +  \frac{\rho^2}{(1-\rho)^2} \leq 4\kappa$.

Thus, using Lemma~\ref{lemma: init bound}, it has
\begin{eqnarray}
\label{bound:distance}
\|\w_{s}^{r} - \w_0^{r}\| &\overset{}{\leq}& \frac{48\sqrt{2n\kappa}}{\lambda\sqrt{m}}\mathcal{O}(\sqrt{nlog(m/\delta)log^2(n/\delta)})
	\overset{}{\leq}\frac{\lambda}{360n\gamma},
\end{eqnarray}
where the last inequality satisfies when the width $m =\Omega(\lambda^{-4}n^{4}\gamma^4 \log^3(n/\delta)) = \Omega(\lambda^{-4}n^{4}\kappa^2log^3(n/\delta))$ with $\gamma = \Theta(\sqrt{\kappa})$ and $\eta\lambda \leq 1/\kappa$.

Now we proceed to determine the upper bound of $\|\mub\|$, which is crucial for bounding $\|\sum_{s=0}^{t-1}\M^{t-s-1}\mub_{s}\|$.
Note that $\|\mub_s\| \leq \|\phib_s\|+\|\psib_s\|$.
Firstly, we derive the bound of $\|\phib_{s}\|$ as:
{{
\begin{eqnarray}
	\| \phib_{s} \| 
	\!{=}\! \sqrt{\sum_{i=1}^n \phib_s[i]^2}
	& \overset{\text{a}}{\leq}&\!\!\!\! \big[\sum_{i=1}^n \big( \frac{ \underset{j\in [n]}{\sup}\!|S_j^{\perp}|\!\sqrt{n}\eta}{m}\big(\!(\!2\!+\!4\beta\!)\|\xib_s\|\!\!+\!\! 2\beta\|\!\xib_{s\!-\!1} \!\|\!\!+\!\!2\!\!\sum_{k=0}^{s-1}\!\beta^{s+1\!-\!k}\|\xib_k\|)\big)^2\big]^{\frac{1}{2}} \nonumber \\
	&\overset{\text{b}}{\leq}&\!\!\!\! 4\eta n R\big((2+4\beta)\|\xib_s\| + 2\beta\|\xib_{s-1} \| + 2\sum_{k=0}^{s-1}\beta^{s+1-k}\|\xib_k\|\big) \nonumber\\
	&\overset{\text{c}}{\leq}&\!\!\!\! 8\eta nR\gamma\|\z_0\|\big((2+4\beta)b^s+  2\beta \rho^{s-1} + \frac{2\rho^{s+3}}{1-\rho})\nonumber\\
	&\overset{\text{d}}{\leq}&\!\!\!\! 48\sqrt{\kappa}\eta n R \gamma \rho^s \|\z_0\| \nonumber \\
	\label{bound: phib}
	&\overset{\text{e}}{\leq}&\!\!\!\! \frac{2\rho^s}{ 15 \sqrt{\kappa}} \|\z_0\|,
\end{eqnarray}}}
where (a) uses Lemma 1 to provide the bound of $|\phib_s[i]|$,\\
(b) uses Lemma~\ref{lemma: bound of S_i} to give the upper bound of $\sup_{j \in [n]}|S_j^{\perp}|$ with $\sup_{j \in [n]}|S_j^{\perp}| \leq 4mR$,\\
(c) uses the inductive hypothesis,\\
(d) uses $\beta < \rho^2$ and $1+2\rho^2 + \rho + \frac{\rho^3}{1-\rho} \leq 3\sqrt{\kappa}$,\\
(e) uses $R\leq \lambda/(360n\gamma)$.

For bounding $\|\psib_s\|$, we have
\begin{eqnarray}
	\|\psib_s\| &=& \|\beta\eta(\H_{s-1} - \H_0)\xib_{s-1} -(1+\beta)\eta(\H_s - \H_0)\xib_s\| \nonumber \\
	&\leq& \beta \eta \|\H_{s-1} - \H_0\|\|\xib_{s-1}\| +(1+\beta)\eta\|\H_s - \H_0\|\|\xib_s\| \nonumber \\
	\!&\overset{\text{a}}{\leq}& \frac{1+\beta + \rho}{90\kappa }\rho^s \|\z_0\| \nonumber\\
	\label{bound: psib}
	 \!&\overset{\text{b}}{\leq}& \frac{\rho^s}{30 \kappa }  \|\z_0\|
\end{eqnarray}
where (a) applies $\beta < \rho^2$, the inductive hypothesis, Lemma \ref{lemma: H_t and H_0} and~(\ref{bound:distance}),\\
(b) uses $\beta < 1$ and $\rho < 1$.

Combines (\ref{bound: phib}) and (\ref{bound: psib}), it has
\begin{eqnarray}
\|\mub_s\| \leq \|\phib_s\|+\|\psib_s\|
\leq (\frac{2}{15\sqrt{\kappa}} + \frac{1}{30\kappa }) \rho^s \|\xib_0\|.
\end{eqnarray}

As a result, we have
\begin{eqnarray}
	\|\sum_{s=0}^{t-1}M^{t-s-1}\mub_s\| \leq \sum_{s=0}^{t-1}\alpha^{t-s-1}\gamma\|\mub_s\|
	&\leq& (\frac{2}{15\sqrt{\kappa}} + \frac{1}{30\kappa })\gamma\|\z_0\|\sum_{s=0}^{t-1}a^{t-s-1}\rho^s \nonumber \\ 
	&\overset{\text{a}}{\leq}&\frac{1}{6\sqrt{\kappa}}\gamma\|\z_0\|6\sqrt{\kappa}\rho^t \nonumber \\
	\label{bound:part_2}
	&\overset{\text{b}}{\leq}& \rho^t\gamma\|\z_0\|
\end{eqnarray}
(a) uses $\sum_{s=0}^{t-1}\alpha^{t-s-1}\rho^s \leq \rho^{t-1}\sum_{s=0}^{t-1}{(\frac{\alpha}{\rho}})^{t-s-1}\leq 6\sqrt{\kappa}\rho^t$.
Finally, by combining (\ref{bound: M_z}) and (\ref{bound:part_2}), it has
\begin{eqnarray}
\|\z_{t}\| &\leq& \|\M^t\z_{0}\| + \|\sum_{s=0}^{t-1}\M^{t-s-1}\mub_{s}\|  \leq \rho^t\gamma\|\z_0\| + \rho^t\gamma \|\z_0\| \leq  \rho^t2\gamma\|\z_0\|,
\end{eqnarray}
which completes the proof.
\end{proof}
\bibliographystyle{elsarticle-num}
 \bibliography{NAG}

\end{document}